%% file: iclr2027_conference.tex
\documentclass{article} 
\usepackage{iclr2027_conference,times}

\input{math_commands.tex}

\usepackage{graphicx}
\usepackage{amsmath}
\usepackage{amssymb}
\usepackage{booktabs}
\usepackage{multirow}
\usepackage{tabularx}
\usepackage[table,dvipsnames]{xcolor}
\usepackage{algorithm}
\usepackage{algorithmic}
\usepackage{enumitem}
\usepackage[most]{tcolorbox}
\usepackage{listings}
\usepackage{hyperref}
\usepackage{url}
\definecolor{oursbg}{RGB}{230,231,250}
\newcommand{\posdelta}[1]{{\scriptsize\textcolor{green!50!black}{+#1}}}
\newcommand{\negdelta}[1]{{\scriptsize\textcolor{red}{-#1}}}

\usepackage{titletoc}

\newcommand{\method}{GraphSkillEvo}

\newcommand{\needappref}[1]{\textcolor{red}{[\textit{appendix ref needed}]}}

\makeatletter
\newenvironment{breakablealgorithm}
  {
    \par\medskip
    \begingroup
    \refstepcounter{algorithm}
    \noindent\hrule height .8pt\kern 2pt
    \renewcommand{\caption}[2][\relax]{%
      {\noindent\textbf{Algorithm \thealgorithm.} ##2\par}%
      \ifx\relax##1\relax
        \addcontentsline{loa}{algorithm}{\protect\numberline{\thealgorithm}##2}%
      \else
        \addcontentsline{loa}{algorithm}{\protect\numberline{\thealgorithm}##1}%
      \fi
      \kern 2pt\hrule\kern 2pt
    }%
  }
  {
    \kern 2pt\hrule
    \par\medskip
    \endgroup
  }
\makeatother

\newtcolorbox{skillbox}{
  enhanced,
  breakable,
  colback=white,
  colframe=black,
  boxrule=0.5pt,
  arc=0pt,
  outer arc=0pt,
  left=6pt,
  right=6pt,
  top=6pt,
  bottom=6pt,
  left skip=0.04\linewidth,
  right skip=0.04\linewidth,
  before skip=8pt,
  after skip=8pt
}

\newtcolorbox{dialogbox}[1][]{
  arc=4mm,
  colback=lightgray!20,
  colframe=blue!30!black,
  rounded corners,
  boxrule=1.5pt,
  fonttitle=\sffamily\bfseries\small,
  coltitle=white,
  toptitle=2mm,
  bottomtitle=2mm,
  title=#1,
  frame style={dashed},
  breakable,
}

\newtcolorbox{dialogbox2}[1][]{
  arc=4mm,
  colback=lightgray!20,
  colframe=green!25!black,
  rounded corners,
  boxrule=1.5pt,
  fonttitle=\sffamily\bfseries\small,
  coltitle=white,
  toptitle=2mm,
  bottomtitle=2mm,
  title=#1,
  frame style={dashed},
  breakable,
}

\title{GraphSkillEvo: Evolutionary Optimization of Graph-Structured Agent Skills}

\author{
Rui Sun\textsuperscript{1}\thanks{Equal contribution}  ,
Zhi Zheng\textsuperscript{2}\textsuperscript{*},
Zhenkun Wang\textsuperscript{3},
Zhichao Lu\textsuperscript{1} \\
\textsuperscript{1}City University of Hong Kong \quad
\textsuperscript{2}National University of Singapore \\
\textsuperscript{3}Southern University of Science and Technology \\
\texttt{rui.sun@my.cityu.edu.hk, zhi.zheng@u.nus.edu} \\
\texttt{wangzhenkun90@gmail.com, zhichao.lu@cityu.edu.hk} \\
}

\iclrfinalcopy 
\begin{document}

\maketitle

\begin{abstract}

Skills can improve the performance of Large Language Model (LLM) agents by providing task-specific procedural guidance, while skill optimization further improves their effectiveness through iterative refinement. However, existing skill optimization methods typically represent skills as unstructured natural-language instructions, creating two key challenges: \textbf{1)} Unstructured skills often lack explicit workflow-level guidance and contain substantial redundancy, making them difficult for LLMs to execute; \textbf{2)} the vast search space of unconstrained natural-language skills makes skill optimization ineffective.
To address these challenges, we propose representing skills as \textbf{graph-structured natural-language artifacts}. In graph-structured skills, each node represents an execution step together with its operational guidance, while directed edges encode context-dependent transitions between steps. Compared to unstructured skills, graph-structured skills can provide clear workflow-level guidance. Moreover, the proposed graph-structured skill can also facilitate skill optimization.
Building on this structured representation, we introduce \textbf{GraphSkillEvo}, a population-based evolutionary optimization framework with mutation and crossover operators for graph-structured skills. By maintaining multiple candidate skills and combining effective components, GraphSkillEvo enables broader and more comprehensive exploration of the structured skill space than purely LLM-based iterative self-refinement.
Extensive experiments across five agent benchmarks demonstrate that GraphSkillEvo consistently outperforms the strong skill optimization baseline SkillOpt, improving average accuracy by 4.01\% on GPT-5.4-nano and 1.76\% on GPT-5.4.
Our code is available at \url{https://github.com/ruisun7/GraphSkillEvo}. 
\end{abstract}


\begin{figure}[H]
    \centering\vspace{-12pt}
    \includegraphics[width=\hsize]{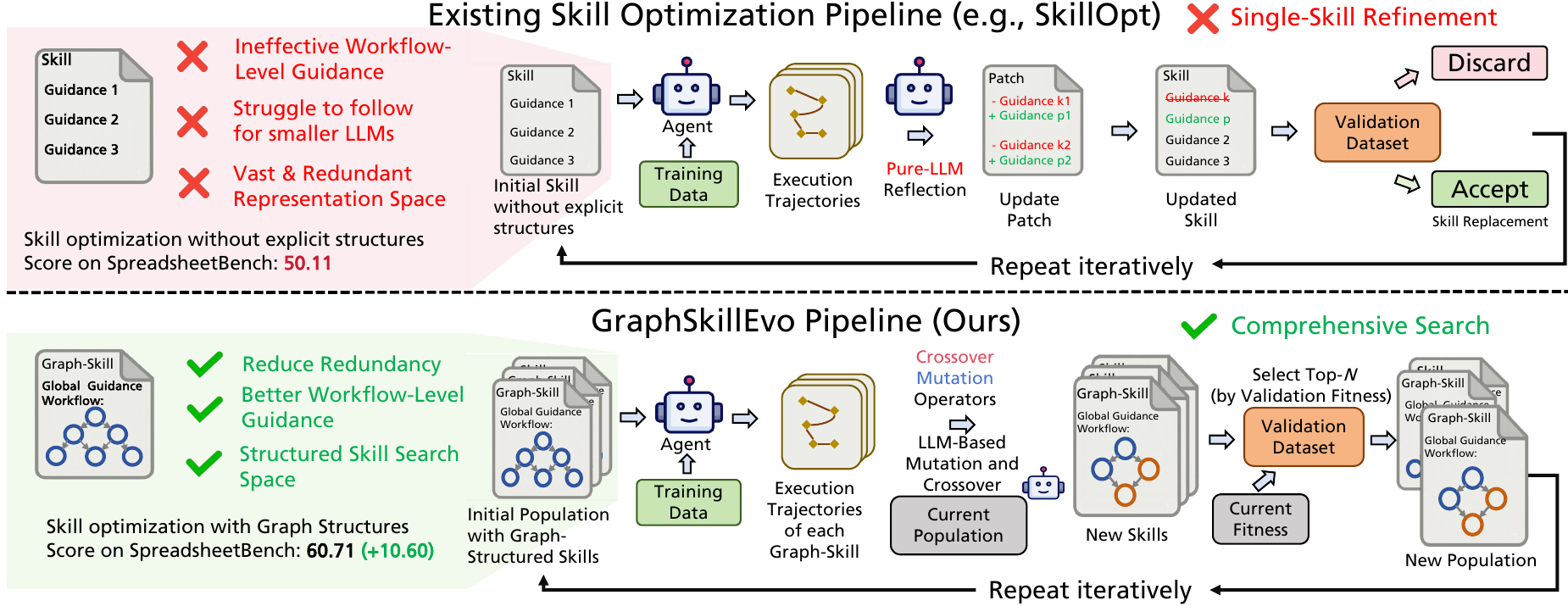}\vspace{-7pt}
    \caption{
    Existing skill optimization methods (e.g., SkillOpt) maintain unstructured skills and optimize skills purely with LLM self-reflection, resulting in suboptimal performance. GraphSkillEvo solves this with graph-structured skills for giving more workflow-level guidance and providing a reduced search space for comprehensive population-based evolutionary search.
    }\vspace{-3pt}
    \label{fig:1}
\end{figure}

\section{Introduction}

Large language models (LLMs) are increasingly deployed as agents across a wide range of real-world applications \citep{schick2023toolformer,wang2023voyager,yang2024swe,yao2022react}. In these agentic settings, skills serve as reusable prompt-level natural-language artifacts that provide task-specific procedural guidance, encoding workflows, domain knowledge, operational rules, and output constraints to help agents complete complex tasks \citep{li2026skillsbench,jiang2026sok}. Beyond improving the performance of a particular agent, an important advantage of skills is that the procedural knowledge they encode can be reused across different LLMs. This cross-model portability is especially valuable as LLMs are rapidly updated and replaced in practice, allowing task-specific capabilities to be preserved without rebuilding the underlying procedures for every newly released model \citep{yang2025qwen3,guo2025deepseek,geminiteam2025gemini25}.

However, manually written or one-shot LLM-generated skills can be incomplete and fragile, motivating recent work on \emph{skill optimization}
\citep{ni2026trace2skill,alzubi2026evoskill,yang2026autoskill,zhang2026coevoskills,wang2026skillx,liu2026skillforge,ma2026skillclaw}.
Existing skill optimization methods (e.g., SkillOpt \citep{yang2026skillopt} shown in Figure \ref{fig:1}), however, typically represent skills as unstructured natural-language instructions without an explicit structure.
This unstructured representation creates two fundamental challenges:
\begin{enumerate}[leftmargin=*]
    \item \textbf{Difficulty in skill execution.}
    Optimized skills often take the form of lengthy checklists or bullet-point instructions that provide only coarse-grained workflow guidance, making it difficult for LLM agents to determine which guidance is relevant at the current stage and what step should follow next. This issue is particularly severe for less capable LLMs (e.g., GPT-5.4-nano), which are more likely to struggle with overlong instructions.

    \item \textbf{Difficulty in skill optimization.}
    The lack of explicit structure also results in a large and redundant search space for skill optimization. Similar workflows can be expressed through many different unstructured textual realizations. As a result, the optimizer must explore many representational variants that do not correspond to meaningful procedural changes.
\end{enumerate}

To address these limitations, as shown in Figure~\ref{fig:1}, motivated by the close analogy between skills and flow diagrams in providing stepwise procedural guidance, we formulate each skill as a \textbf{graph-structured natural-language artifact}. In a graph-structured skill, each node represents an agentic execution step and contains self-contained operational guidance, including the instructions, rules, and constraints relevant to that step. Each directed edge represents a context-dependent transition between execution steps, allowing different task conditions to induce different execution paths through the graph. For \textbf{skill execution}, this formulation makes the underlying workflow explicit and helps agents identify the guidance relevant to each execution step. For \textbf{skill optimization}, the graph structure reduces representational redundancy by explicitly organizing execution steps and their dependencies, thereby providing a more compact and structured search space than unstructured natural-language skills.


Building on this structured search space, we introduce \textbf{\method{}}, a population-based \textbf{evolutionary computation (EC) framework} for optimizing graph-structured agent skills.
\method{} maintains a population of candidate skills to preserve the diversity of high-quality graph-structured skills throughout optimization and evolves them using structure-aware mutation and crossover operators.
Mutation revises individual skills based on their execution trajectories, while crossover recombines complementary and effective graph components from different candidates.
Together, these mechanisms enable broader exploration beyond purely LLM-based iterative self-refinement and facilitate the discovery of higher-quality skills. Our contributions are as follows:
\begin{enumerate}[leftmargin=*]
    \item We formulate agent skills as graph-structured natural-language artifacts that explicitly represent execution steps and context-dependent transitions, providing clearer workflow-level guidance and reducing representational redundancy.
    \item We introduce \method{}, a population-based evolutionary computation framework with structure-aware mutation and crossover operators for effectively exploring and optimizing graph-structured skills.
    \item We conduct extensive experiments across diverse agent benchmarks, demonstrating that \method{} consistently outperforms strong skill-optimization baselines across two different LLMs, two different harnesses, and five benchmark settings.
\end{enumerate}




\section{Preliminary}
\subsection{Problem Definition: Skill Optimization}

Let $\mathcal{A}$ denote an agent composed of an LLM and its execution harness \citep{guo2026question}. For a given task, a skill $s$ is a natural-language artifact supplied to the agent during execution to help solve instances of that task. 
Such skills can be manually written, generated by LLMs in one shot, or further refined through skill optimization~\citep{ni2026trace2skill}.
For a task instance $x$, execution with the skill $s$ produces a trajectory $\tau_{x,s}$ and a score $r_{x,s}$ computed by a task-specific scoring function $R$:
\[
\tau_{x,s}=\mathcal{A}(x,s),
\qquad
r_{x,s}=R\bigl(x,\tau_{x,s}\bigr).
\]

The optimization goal is to find a skill that maximizes the performance of the agent on the task:
\[
s^\star \in \arg\max_s J_{D}(s),
\qquad
J_{D}(s)
=
\frac{1}{|D|}
\sum_{x \in D} r_{x,s}.
\]
Here, $D$ is a dataset of instances of the task, and
$J_{D}(s)$ measures task performance of the agent using skill $s$.
Following existing skill optimization methods~\citep{yang2026skillopt}, we optimize only the skill artifact while keeping the LLM parameters and execution harness fixed.
In practice, $D$ consists of three subsets: $D_{\mathrm{train}}$, $D_{\mathrm{val}}$, and $D_{\mathrm{test}}$.
The training set $D_{\mathrm{train}}$ is used to collect execution trajectories, which provide feedback for proposing new skills.
The validation set $D_{\mathrm{val}}$ is used to assess candidate skills during optimization.
The test set $D_{\mathrm{test}}$ is used only for final evaluation.



\subsection{Skill Optimization Method}
Manually written skills or skills generated by LLMs in one shot are usually incomplete and fragile. So, recent methods automatically construct or distill skills from execution trajectories and interaction experience (e.g., EvoSkill~\citep{alzubi2026evoskill}, Trace2Skill~\citep{ni2026trace2skill}, SkillX~\citep{wang2026skillx}, SkillOpt~\citep{yang2026skillopt}).
EvoSkill discovers and refines skills through iterative failure analysis and Pareto-based selection \citep{alzubi2026evoskill}.
Trace2Skill consolidates multiple trajectory patches into a single portable skill via parallel merging \citep{ni2026trace2skill}.
SkillX extracts multi-level skills from execution trajectories, and constructs a skill library via iterative refinement and exploratory skill expansion \citep{wang2026skillx}. As a representative skill optimization method illustrated in Figure~\ref{fig:1}, SkillOpt \citep{yang2026skillopt} starts from an initial skill \(s^{(0)}\). At iteration \(t\), the current skill \(s^{(t)}\) is provided to agent \(\mathcal{A}\) and executed on the training set \(D_{\mathrm{train}}\), producing execution trajectories as $\mathcal{T}^{(t)}
=
\{\tau_{x,s^{(t)}}\mid x\in D_{\mathrm{train}}\}.$
An agent $\mathcal{A}^{\mathrm{patch}}$ analyzes these trajectories
to generate skill update patches, which are applied to the current
skill to obtain a candidate skill:
\[
\mathrm{Patch}^{(t)} = \mathcal{A}^{\mathrm{patch}}\left(s^{(t)}, \mathcal{T}^{(t)}\right), \qquad \tilde{s}^{(t+1)} = s^{(t)} \oplus \mathrm{Patch}^{(t)},
\
\]
where $\oplus$ denotes textual patch application.
Finally, the candidate skill is evaluated on \(D_{\mathrm{val}}\).
If \(\tilde{s}^{(t+1)}\) improves validation performance over \(s^{(t)}\), then \(s^{(t+1)}\) is set to \(\tilde{s}^{(t+1)}\); otherwise, \(s^{(t+1)}\) remains \(s^{(t)}\).
This process is repeated for multiple rounds.

Though demonstrating solid refinements, existing skill optimization methods still have two main limitations. \textbf{1)} First, existing methods typically optimize skills as unconstrained natural-language artifacts without explicit structural constraints. As a result, optimized skills can become lengthy and redundant while providing limited workflow-level guidance. GraphSkillEvo instead represents skills as graph-structured natural-language artifacts, making the execution workflow explicit. \textbf{2)} Second, existing methods optimize skills in a large and redundant search space. GraphSkillEvo, instead, provides a more compact and well-structured search space and evolves a population of skills through mutation and crossover operators. This results in a more comprehensive exploration compared to the pure LLM-based self-refinement in existing skill optimization methods.

\section{Methodology: GraphSkillEvo}

\subsection{Graph-Structured Skills}

To address the challenges of coarse workflow-level guidance and redundant search space faced by existing methods in optimizing unstructured agent skills, this paper formulates skills as graph-structured natural-language artifacts.
Formally, a graph-structured skill $s=\langle h_s,g_s\rangle$
comprises global guidance $h_s$ and a directed graph
$g_s=(V_s,E_s)$ with node set $V_s$ and edge set $E_s$.

\textbf{(1) Global Guidance $h_s$.}
A skill may include task descriptions, general principles, and shared execution templates that apply across execution steps and are not specific to any individual node.
The global guidance $h_s$ collects these shared instructions.

\textbf{(2) Node Set $V_s$.}
The node set $V_s=\{v_1,\ldots,v_{n_s}\}$ contains $n_s$
reusable nodes. Each node describes an execution step, such as
parsing the task goal, retrieving evidence, performing an operation,
or verifying the final answer, together with the instructions,
rules, and constraints required to perform that step. 

\textbf{(3) Edge Set $E_s$.}
The edge set is specified through $M$ workflows
$\{(c_m,p_m)\}_{m=1}^{M}$.
Each workflow addresses a particular situation within the task,
with $c_m$ specifying its applicability condition.
The execution path $p_m$ is an ordered sequence of nodes from $V_s$
that specifies the order in which the agent follows the corresponding execution steps.
Consecutive nodes in each path define directed edges in $E_s$,
representing transitions between execution steps under the corresponding workflow’s applicability condition.
Different workflows may share nodes while prescribing
different execution paths.


\textbf{Example.} 
We illustrate with an example skill:
\begin{skillbox}
\small

\textbf{Name:} Household Skill

\textbf{Global Guidance $h_s$:}

\textit{Task Description:}
Complete household tasks by navigating rooms, interacting with objects,
and using appliances.

\textit{General Principles:}
Never repeat the same action more than twice in a row. If stuck, move to
a different unexplored location.
Always pick an action from the admissible action list. Do not invent actions.

\textit{\ldots}

\textbf{Node Set $V_s$:}

\textit{Explore Object:}
Systematically search surfaces and containers, opening closed containers
before determining that the target object is absent.

\textit{Take Object:}
When the required object is visible and reachable, take it immediately
before moving elsewhere.

\textit{\ldots}

\textbf{Edge Set $E_s$} (specified through workflows) \textbf{:}

\textit{Pick \& Place:}
\begin{itemize}[leftmargin=*]
\item Use When: Put one instance of the requested object in/on the requested receptacle.
\item Workflow: 1. Parse Goal; 2. Explore Object; 3. Take Object; 4. Find Destination Receptacle; 5. Place Object; 6.  Verify Completion;
\end{itemize}
\textit{\ldots}
\end{skillbox}

Graph-structured skills have several attractive properties for large language model agents.

\begin{enumerate}[leftmargin=*]
\item \textbf{Low Redundancy.} Instructions shared by multiple workflows can be specified once in a reusable node and incorporated into multiple execution paths, rather than being repeated across different sections of a skill document.

\item \textbf{Explicit workflow guidance.} Each workflow specifies an execution path consisting of an ordered
sequence of execution steps, where each step corresponds to a node that contains the instructions, rules, and constraints required for that step. This structure helps LLM agents follow a suitable and precise stepwise workflow and to determine which instruction is relevant at each stage of execution.

\item \textbf{Providing a more compact and structured search space for GraphSkillEvo.}
Instead of separately searching over many unstructured textual realizations of similar workflows, the optimizer can directly operate on explicit execution steps and their dependencies.

\end{enumerate}

\begin{figure*}
    \centering
    \includegraphics[width=\hsize]{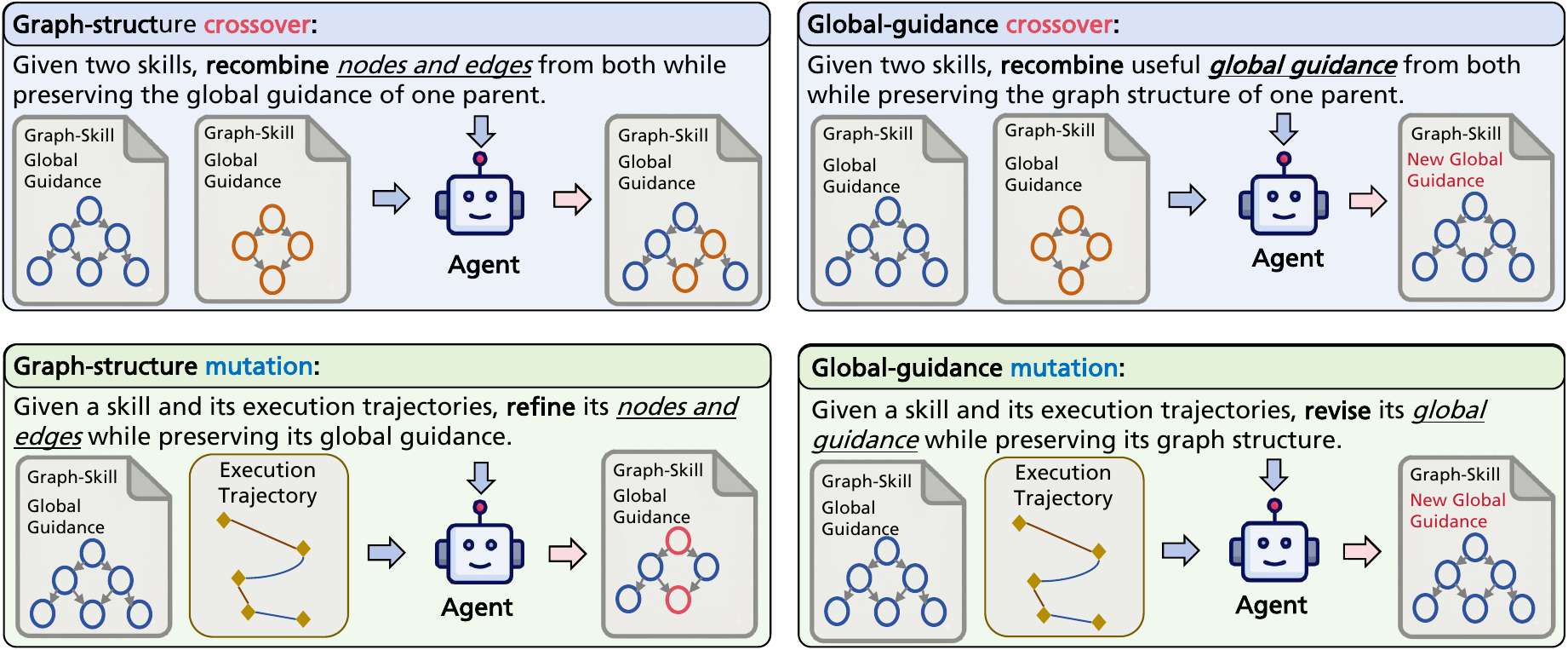}
    \caption{Four evolutionary operators used in GraphSkillEvo.
Global-guidance mutation revises the global guidance of a parent
skill based on its execution trajectories, while graph-structure
mutation updates its nodes and edges by refining node instructions,
adding or removing nodes, and adjusting execution paths.
Global-guidance crossover recombines useful global guidance from
two parent skills while preserving one parent's graph structure,
and graph-structure crossover recombines nodes and edges from
two parent skills while preserving one parent's global guidance.
    }
    \label{fig:3}
\end{figure*}

\subsection{Evolutionary Optimization over Graph-Structured Skills}

To more comprehensively explore the structured skill space, we introduce GraphSkillEvo, a population-based evolutionary optimization framework with mutation and crossover operators for graph-structured skills.
The population preserves multiple high-quality skills throughout optimization. 
Mutation modifies an individual skill by incorporating feedback from its execution trajectories, while crossover transfers beneficial components between graph-structured skills.



The overall optimization procedure of \method{} consists of the following four steps.

\textbf{Step 0: Population initialization.}
\method{} initializes a population $\mathcal{P}^{(1)}=\{s^{(1)}_i\}_{i=1}^{N}$ containing $N$ skills. In addition to the initial skill, the remaining skills are generated by an initialization prompt that provides the task context and asks the LLM to create diverse graph-structured skills. After initialization, every individual $s^{(1)}_i\in\mathcal{P}^{(1)}$ is evaluated on the full validation dataset $D_{\mathrm{val}}$, and its validation score $J_{D_{val}}(s^{(1)}_i)$ is used as its fitness value.

\textbf{Step 1: Execution on the training set.}
At each generation $t$, \method{} samples a small batch of $B$ instances
$\mathcal{B}^{(t)}$ from training dataset $D_{\mathrm{train}}$. Each skill $s^{(t)}_i\in\mathcal{P}^{(t)}$ in the current
population is attached to the LLM agent \(\mathcal{A}\) and executed on these instances, producing execution trajectories
\[
\mathcal{T}_i^{(t)}
=
\{\tau_{x,s_i^{(t)}}\mid x\in \mathcal{B}^{(t)}\}.
\]
For each skill, let $\mathcal{F}_i^{(t)}$ denote the failed trajectories
retained as reflection information for generating new skills. Formally,
\[
\mathcal{F}_i^{(t)}\subseteq
\left\{\tau_{x,s_i^{(t)}}\;\middle|\;
x\in\mathcal{B}^{(t)},\ r_{x,s_i^{(t)}}=0\right\},
\qquad
\left|\mathcal{F}_i^{(t)}\right|\le K.
\]
Here, $r_{x,s_i^{(t)}}=0$ indicates that executing skill
$s_i^{(t)}$ on instance $x$ fails, and $K$ is the maximum number of failed trajectories retained for each skill.

\textbf{Step 2: Generation of new skills.}
\method{} generates $N$ new skills. Each new skill is generated through the following three substeps:




\begin{enumerate}[leftmargin=*]
    \item Step 2.1: Operator selection.
    \method{} selects an operator from four operators using a round-robin schedule.

    \item Step 2.2: Skill selection.
    \method{} selects parent skill(s) from the current population to generate the new skill. The selection probability is $p \propto 1/(r+N)$, where $r$ denotes the fitness rank of the corresponding skill within the population and $N$ is the population size.

    \item Step 2.3: Skill generation.
    For each newly generated skill $\widetilde{s}_j^{(t+1)}$, the operator and parent skill(s) used to generate
    $\widetilde{s}_j^{(t+1)}$ are denoted by $o_j^{(t)}$ and
    $s_{\mathrm{parent},j}^{(t)}$, respectively. Here,
    $s_{\mathrm{parent},j}^{(t)}$ represents one parent for mutation and two
    parents for crossover, while
    $\mathcal{F}_{\mathrm{parent},j}^{(t)}$ denotes the associated retained failed
    execution trajectories. 
    The LLM agent $\mathcal{A}^{\mathrm{gen}}$ generates each new skill $\widetilde{s}_j^{(t+1)}$ using the selected operator and parent skill(s), with the corresponding failed trajectories provided only for mutation.

    $$
    \tilde{s}_j^{(t+1)}
    =
    \begin{cases}
    \mathcal{A}^{\mathrm{gen}}\!\left(
    o_j^{(t)}, s_{\mathrm{parent},j}^{(t)}, \mathcal{F}_{\mathrm{parent},j}^{(t)}
    \right), & \text{if } o_j^{(t)} \text{ is a mutation operator},\\[4pt]
    \mathcal{A}^{\mathrm{gen}}\!\left(
    o_j^{(t)}, s_{\mathrm{parent},j}^{(t)}
    \right), & \text{if } o_j^{(t)} \text{ is a crossover operator},
    \end{cases}
    \qquad j=1,\ldots,N.
    $$

\end{enumerate}

As shown in Figure~\ref{fig:3}, \method{} uses the following four evolutionary operators:
\begin{enumerate}[leftmargin=*]
    \item \textbf{Global-guidance mutation} revises the global guidance of a selected skill using LLM-based self-reflection.
    \item \textbf{Graph-structure mutation} revises the nodes and edges of a selected skill using its reflection information. The revisions include refining node instructions, adding or deleting reusable nodes, and adjusting task workflows.
    \item \textbf{Global-guidance crossover} recombines useful global guidance from two selected skills while preserving the graph structure of one of them.
    \item \textbf{Graph-structure crossover} recombines useful nodes and edges from two selected skills while preserving the global guidance of one of them.
\end{enumerate}
Detailed prompts for these operators are provided in Appendix~\ref{app:operator_prompts}.

\textbf{Step 3: Population selection.}
After generating the $N$ new skills, \method{} evaluates each of them on the full validation set $D_{\mathrm{val}}$. It then updates the population by retaining the $N$ skills with the highest fitness values among the current population and the newly generated skills. Let $\widetilde{\mathcal{P}}^{(t+1)}=\{\widetilde{s}_j^{(t+1)}\}_{j=1}^{N}$ denote the set of newly generated skills. The next-generation population is:
\[
\mathcal{P}^{(t+1)}
\in
\underset{\substack{\mathcal{S}\subseteq
\mathcal{P}^{(t)}\cup\widetilde{\mathcal{P}}^{(t+1)}\\
|\mathcal{S}|=N}}{\arg\max}
\sum_{s\in\mathcal{S}} J_{D_{\mathrm{val}}}(s).
\]
Steps~1-3 are repeated for $T$ generations, after which the skill with the highest fitness value in the final population is returned as the optimized graph-structured skill.

\vspace{-5pt}
\section{Experiments}
\vspace{-2pt}
\label{sec:experiments}

\textbf{Benchmarks.}
We evaluate \method{} on five benchmarks: SearchQA \citep{dunn2017searchqa}, SpreadsheetBench \citep{ma2024spreadsheetbench} (abbreviated as Spreadsheet in tables), DocVQA \citep{mathew2021docvqa}, LiveMathematicianBench \citep{he2026livemathematicianbench} (abbreviated as LiveMath), and ALFWorld \citep{shridhar2020alfworld}.
These benchmarks cover fact-based question answering, spreadsheet manipulation, visual document understanding, mathematical multiple-choice reasoning, and embodied interaction.
For each benchmark, we divide the data into a training set, a validation set, and a test set.
The details of each benchmark are provided in Appendix~\ref{app:benchmarks} and Appendix \ref{app:dataset_splits}.

\textbf{Metrics.}
We report the average success rate on the test set.
For SearchQA, DocVQA, and LiveMath, correctness is measured by exact match accuracy.
For SpreadsheetBench, a task is correct only when the workbook matches the gold answer at all required locations across all evaluation cases.
For ALFWorld, correctness is measured by the pass rate within an interaction limit.

\textbf{Baselines.}
We compare against four skill sources. \textbf{1)} \textit{No skill} runs the benchmark without any skill. \textbf{2)} \textit{Human skill} uses a skill written by an expert. \textbf{3)} \textit{LLM skill} uses a skill generated by an LLM from the task description. \textbf{4)} \textit{SkillOpt} iteratively optimizes skills using rollout reflections, selected edits, and validation gating. The implementation details of the baselines are provided in Appendix \ref{app:baseline_detail}.

\textbf{LLMs.}
All experiments use GPT-5.4 \citep{openai2026gpt54} and GPT-5.4-nano. We use medium reasoning effort for GPT-5.4 and GPT-5.4-nano. In each experimental setting, the same LLM is used for task execution and skill optimization in both GraphSkillEvo and SkillOpt.

\textbf{Harness.}
We evaluate \method{} both without an agent harness and with the Codex harness.
Without a harness, the skill is incorporated into the model instructions for each benchmark.
With the Codex harness, Codex is invoked through its software development kit (SDK), and each task is assigned a separate local workspace.
Each workspace contains the task description, any associated input files, and the skill.
Codex is instructed to read the skill and follow its guidance while solving the task.
Codex operates in the \texttt{workspace-write} sandbox with interactive approvals disabled.
We leave the ALFWorld cells blank for the Codex harness because ALFWorld requires a persistent environment interaction, which is not supported by the standard Codex adapter.

\textbf{Optimization parameters.}
During evolution, we set the population size to $N=4$ and run $T=5$ generations. At each generation, \method{} samples 15 instances from $D_{\mathrm{train}}$ for execution, and uses up to 5 failed instances to build the reflection information for mutation. The four operators are selected in a round-robin schedule.

\begin{table*}[t!]
\centering
\footnotesize
\vspace{-10pt}

\caption{Main results across five benchmarks, two LLMs, and two agent harnesses. Each entry reports the success rate on the test set. Higher values indicate better performance. Bold numbers mark the best-reported result among all skill sources for the same model and benchmark.}
\resizebox{\textwidth}{!}{
\setlength{\tabcolsep}{3pt}
\renewcommand{\arraystretch}{1.05}
\begin{tabular}{llllllll}
\toprule[1.5pt]

Model & Skill source & SearchQA & Spreadsheet & DocVQA & LiveMath & ALFWorld & Average \\
\midrule

\multicolumn{8}{@{}>{\columncolor{gray!20}[0pt][0pt]}l@{}}{\hspace{\tabcolsep}\textbf{No harness}} \\
\multirow{5}{*}{GPT-5.4}
& No Skill      &77.50 &39.16 &79.05 &33.60 &73.13 &60.49 \\
& Human skill   &77.71\,\posdelta{0.21} &37.85\,\negdelta{1.31} &84.04\,\posdelta{4.99} &31.72\,\negdelta{1.88} &81.84\,\posdelta{8.71} &62.63\,\posdelta{2.14} \\
& LLM skill     &78.19\,\posdelta{0.69} &35.59\,\negdelta{3.57} &88.23\,\posdelta{9.18} &35.21\,\posdelta{1.61} &75.86\,\posdelta{2.73} &62.62\,\posdelta{2.13} \\
& SkillOpt      &82.21\,\posdelta{4.71} &64.87\,\posdelta{25.71} &89.30\,\posdelta{10.25} &47.58\,\posdelta{13.98} & 86.56\,\posdelta{13.43} &74.10\,\posdelta{13.61} \\
& \cellcolor{oursbg}GraphSkillEvo & \cellcolor{oursbg}\textbf{83.80}\,\posdelta{6.30} & \cellcolor{oursbg}\textbf{69.40}\,\posdelta{30.24} & \cellcolor{oursbg}\textbf{90.37}\,\posdelta{11.32} & \cellcolor{oursbg}\textbf{48.65}\,\posdelta{15.05} & \cellcolor{oursbg}\textbf{87.06}\,\posdelta{13.93} & \cellcolor{oursbg}\textbf{75.86}\,\posdelta{15.37} \\
\midrule

\multirow{5}{*}{GPT-5.4-nano}
& No Skill      &58.12 &35.12 &36.72 &23.93 &41.29 &39.04 \\
& Human skill   &62.36\,\posdelta{4.24} &34.64\,\negdelta{0.48} &59.09\,\posdelta{22.37} &25.80\,\posdelta{1.87} &50.25\,\posdelta{8.96} &46.43\,\posdelta{7.39} \\
& LLM skill     &60.14\,\posdelta{2.02} &32.73\,\negdelta{2.39} &66.49\,\posdelta{29.77} &24.19\,\posdelta{0.26} &53.73\,\posdelta{12.44} &47.46\,\posdelta{8.42} \\
& SkillOpt      &69.52\,\posdelta{11.40} &50.11\,\posdelta{14.99} &77.80\,\posdelta{41.08} &\textbf{29.56}\,\posdelta{5.63} &57.46\,\posdelta{16.17} &56.89\,\posdelta{17.85} \\
& \cellcolor{oursbg}GraphSkillEvo &\cellcolor{oursbg}\textbf{72.93}\,\posdelta{14.81} &\cellcolor{oursbg}\textbf{60.71}\,\posdelta{25.59} &\cellcolor{oursbg}\textbf{80.92}\,\posdelta{44.20} &\cellcolor{oursbg}28.76\,\posdelta{4.83} &\cellcolor{oursbg}\textbf{61.19}\,\posdelta{19.90} &\cellcolor{oursbg}\textbf{60.90}\,\posdelta{21.86} \\

\midrule[1.2pt]

\multicolumn{8}{@{}>{\columncolor{gray!20}[0pt][0pt]}l@{}}{\hspace{\tabcolsep}\textbf{Codex harness}} \\
\multirow{5}{*}{GPT-5.4}
& No Skill      &79.42 &56.07 &82.35 &54.03 &- &67.97 \\
& Human skill   &82.36\,\posdelta{2.94} &48.57\,\negdelta{7.50} &86.36\,\posdelta{4.01} &51.61\,\negdelta{2.42} &- &67.23\,\negdelta{0.74} \\
& LLM skill     &81.93\,\posdelta{2.51} &46.79\,\negdelta{9.28} &86.10\,\posdelta{3.75} &54.03\,\posdelta{0.00} &- &67.21\,\negdelta{0.76} \\
& SkillOpt      &83.02\,\posdelta{3.60} &77.14\,\posdelta{21.07} &87.43\,\posdelta{5.08} &60.21\,\posdelta{6.18} &- &76.95\,\posdelta{8.98} \\
& \cellcolor{oursbg}GraphSkillEvo &\cellcolor{oursbg}\textbf{83.26}\,\posdelta{3.84} &\cellcolor{oursbg}\textbf{79.28}\,\posdelta{23.21} &\cellcolor{oursbg}\textbf{89.30}\,\posdelta{6.95} &\cellcolor{oursbg}\textbf{61.29}\,\posdelta{7.26} &\cellcolor{oursbg}- &\cellcolor{oursbg}\textbf{78.28}\,\posdelta{10.31} \\

\bottomrule[1.5pt]
\end{tabular}}

\label{tab:main_results}
\end{table*}

\begin{figure}[t]
\centering
\begin{minipage}[t]{0.32\linewidth}
\centering
\includegraphics[width=\linewidth]{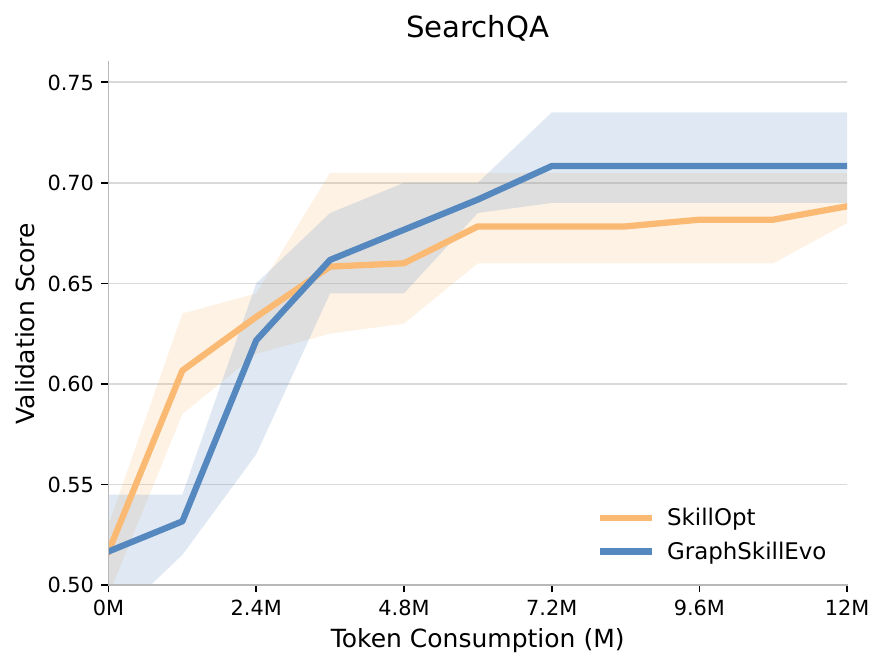}
\end{minipage}\hfill
\begin{minipage}[t]{0.32\linewidth}
\centering
\includegraphics[width=\linewidth]{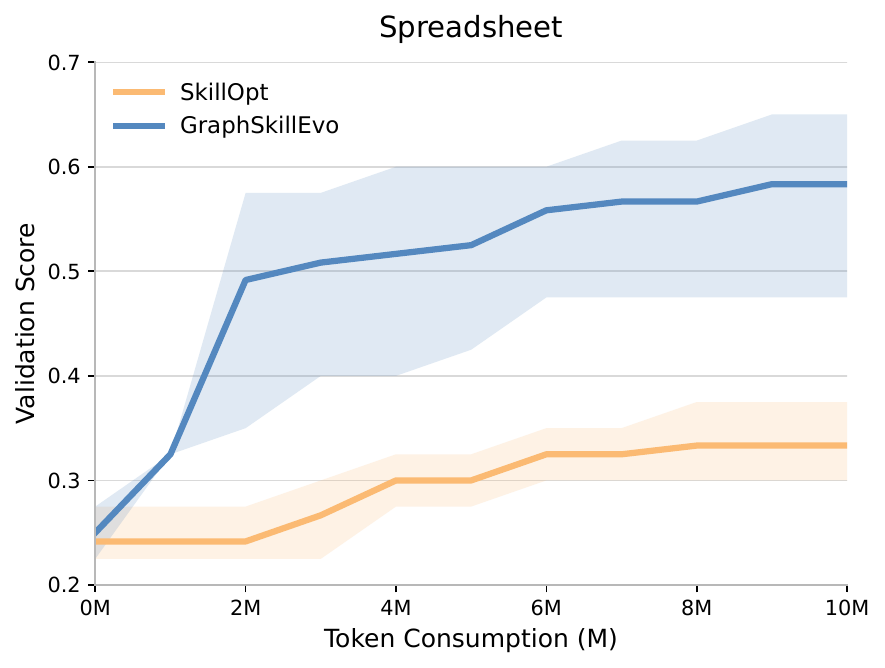}
\end{minipage}\hfill
\begin{minipage}[t]{0.32\linewidth}
\centering
\includegraphics[width=\linewidth]{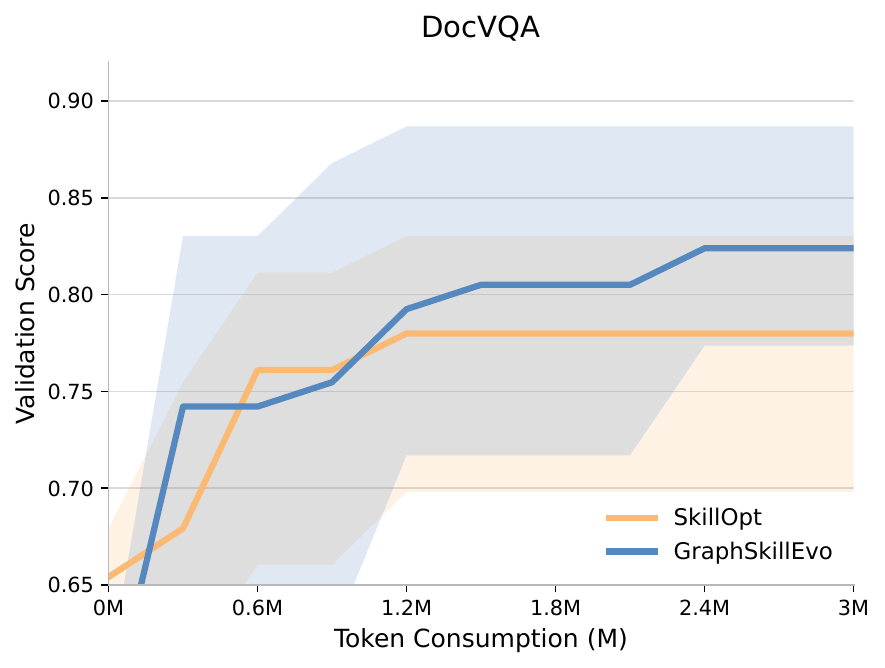}
\end{minipage}
\vspace{-5pt}
\caption{Optimization curve comparison between SkillOpt and \method{}.}
\label{fig:optimization-curves}
\vspace{-5pt}
\end{figure}

\subsection{Main Results}
\label{subsec:main_results}

Table~\ref{tab:main_results} presents the main results across five benchmarks, two LLMs, and two agent harnesses. All reported results are averages over three repeated skill optimization runs.
We compare \method{} with no-skill execution, human-written skills, LLM-generated skills, and SkillOpt. All entries are test-set success rates. Across the 14 model--harness--benchmark settings, \method{} achieves the best result in 13 settings. 
\textbf{1)} Relative to no-skill execution, \method{} improves the average success rate by 15.37\% in the GPT-5.4 no-harness setting, 21.86\% in the GPT-5.4-nano no-harness setting, and 10.31\% in the GPT-5.4 Codex-harness setting.
\textbf{2)} Compared with SkillOpt, a promising skill-optimization method, \method{} achieves average gains of 1.76\% under GPT-5.4 without a harness, 4.01\% under GPT-5.4-nano without a harness, and 1.33\% under GPT-5.4 with the Codex harness.

\textbf{Small and less capable models benefit the most.} Averaged across the five benchmarks, \method{} outperforms SkillOpt by 4.01\% on GPT-5.4-nano, compared with 1.76\% on GPT-5.4. 
Procedural benchmarks see particularly large improvements. \method{} improves over SkillOpt by 10.60\% on SpreadsheetBench and 3.73\% on ALFWorld. These gains suggest that the clear workflow guidance provided by graph-structured skills is especially helpful for tasks that require procedural execution, especially when agents need to interact with an external environment.
The only exception is LiveMath with GPT-5.4-nano, where GraphSkillEvo trails SkillOpt by 0.80\%.

Taken together, these results demonstrate that \method{} is broadly effective across heterogeneous agent tasks, different LLM settings, and agent harnesses. The operator prompts and the generated skills are listed in Appendix~\ref{app:prompts_skills}. 
We also report the significance tests in Appendix~\ref{app:significance_test} and a case study in Appendix~\ref{app:case_study}.


\paragraph{Optimization Curves}
\label{app:optimization_curves}

Figure~\ref{fig:optimization-curves} plots the optimization curves of \method{} and SkillOpt in the GPT-5.4-nano setting without an agent harness.
The validation score is shown against the total number of tokens consumed during optimization.
Each curve is averaged over three experiments. 
SkillOpt shows early convergence on performance, while GraphSkillEvo is able to converge to better performance via continuous performance updates.

\begin{table}[t]
\centering
\vspace{-5pt}
\small
\setlength{\tabcolsep}{5pt}
\renewcommand{\arraystretch}{1.05}

\caption{Token consumption of SkillOpt and \method{} across different models and benchmarks. All values are reported in millions (M).}
\begin{tabularx}{\textwidth}{
    ll
    *{6}{>{\centering\arraybackslash}X}
}
\toprule[1.5pt]

Model & Method & SearchQA & Spreadsheet & DocVQA
      & LiveMath & ALFWorld & Total \\
\midrule

\multirow{2}{*}{GPT-5.4}
    & SkillOpt   & 38.11 M & 11.16 M & 3.69 M & 2.86 M & 25.26 M & 81.08 M \\
    & \cellcolor{oursbg}\method & \cellcolor{oursbg}12.54 M & \cellcolor{oursbg}8.82 M & \cellcolor{oursbg}4.56 M & \cellcolor{oursbg}3.50 M & \cellcolor{oursbg}32.53 M & \cellcolor{oursbg}\textbf{61.94 M} \\
\midrule

\multirow{2}{*}{GPT-5.4-nano}
    & SkillOpt   & 44.68 M & 25.22 M & 3.67 M & 1.78 M & 28.18 M & 103.54 M \\
    & \cellcolor{oursbg}\method & \cellcolor{oursbg}12.69 M & \cellcolor{oursbg}9.16 M & \cellcolor{oursbg}3.44 M & \cellcolor{oursbg}2.90 M & \cellcolor{oursbg}47.75 M & \cellcolor{oursbg}\textbf{75.94 M} \\

\bottomrule[1.5pt]
\end{tabularx}

\vspace{-5pt}
\label{tab:token-consumption}
\end{table}

\paragraph{Token Consumption}
Table~\ref{tab:token-consumption} reports the token consumption of our method and the SkillOpt baseline, including the total consumption and the consumption on each benchmark.
For the total consumption, SkillOpt uses 1.31 times as many tokens as \method{} with GPT-5.4 and 1.36 times as many tokens with GPT-5.4-nano.
Overall, the totals in Table~\ref{tab:token-consumption} are lower for \method{} than for SkillOpt under both model settings.
These results show that our method achieves stronger performance while using substantially fewer optimization tokens than SkillOpt.

\begin{table*}[t]
\centering
\small
\vspace{-5pt}
\setlength{\tabcolsep}{12pt}
\renewcommand{\arraystretch}{1.05}

\caption{Effect of the graph-structured skill representation across five benchmarks. Each entry reports the test-set success rate, expressed as a percentage. The unstructured counterpart retains global guidance and node-level instructions without explicit workflow organization.}
\begin{tabularx}{\textwidth}{l*{5}{>{\raggedright\arraybackslash}X}}
\toprule[1.5pt]

Skill & SearchQA & Spreadsheet & DocVQA & LiveMath & ALFWorld \\
\midrule
\rowcolor{oursbg}
Graph-structured  &\textbf{72.93} &\textbf{60.71} &\textbf{80.92} &\textbf{28.76} &\textbf{61.19} \\
Unstructured &68.41\,\negdelta{4.52} &58.21\,\negdelta{2.50} &76.73\,\negdelta{4.19} &27.41\,\negdelta{1.35} &60.44\,\negdelta{0.75} \\

\bottomrule[1.5pt]
\end{tabularx}
\vspace{-5pt}

\label{tab:graph}
\end{table*}

\begin{table*}[t]
\centering
\small
\vspace{-5pt}
\setlength{\tabcolsep}{7pt}
\renewcommand{\arraystretch}{1.05}

\caption{Ablation study of the graph structure and evolutionary operators.}
\begin{tabularx}{\textwidth}{
    l
    >{\raggedright\arraybackslash}X
    >{\raggedright\arraybackslash}X
    >{\raggedright\arraybackslash}X
    >{\raggedright\arraybackslash}X
}
\toprule[1.5pt]
Skill & \shortstack{SearchQA} & \shortstack{Spreadsheet} & \shortstack{DocVQA} & \shortstack{Average} \\
\midrule
\rowcolor{oursbg}
\method                      & \textbf{72.93} & \textbf{60.71} & \textbf{80.92} & \textbf{71.52} \\
\midrule
\shortstack[l]{\textit{w/o} graph structure} & 68.83\,\negdelta{4.10} & 50.59\,\negdelta{10.12} & 72.81\,\negdelta{8.11} & 64.08\,\negdelta{7.44} \\
\textit{w/o} mutation        & 57.19\,\negdelta{15.74} & 35.11\,\negdelta{25.60} & 71.20\,\negdelta{9.72} & 54.50\,\negdelta{17.02} \\
\textit{w/o} crossover       & 71.80\,\negdelta{1.13} & 53.56\,\negdelta{7.15} & 74.41\,\negdelta{6.51} & 66.59\,\negdelta{4.93} \\
\bottomrule[1.5pt]

\label{tab:ablation}
\end{tabularx}
\vspace{-10pt}
\end{table*}

\begin{table*}[t]
\centering
\vspace{-5pt}
\small
\setlength{\tabcolsep}{10pt}
\renewcommand{\arraystretch}{1.05}

\caption{Cross-model transferability of optimized skills.
\textbf{Baseline} denotes execution on GPT-5.4 without a skill.
\textbf{Direct} denotes using a skill optimized with GPT-5.4 and then applied
on GPT-5.4, and \textbf{Transferred} denotes using a skill optimized with
GPT-5.4-nano and then applied on GPT-5.4.}
\label{tab:transfer-performance}
\begin{tabularx}{\textwidth}{
    >{\raggedright\arraybackslash}X
    >{\raggedright\arraybackslash}X
    >{\raggedright\arraybackslash}X
    >{\raggedright\arraybackslash}X
    >{\raggedright\arraybackslash}X
}
\toprule[1.5pt]
Benchmark & Method & Baseline & Direct & Transferred \\
\midrule

\multirow{2}{*}{SearchQA}
& SkillOpt & 77.50 & 82.21\,\posdelta{4.71} & 83.92\,\posdelta{6.42} \\
& \cellcolor{oursbg}\method{} & \cellcolor{oursbg}77.50 & \cellcolor{oursbg}83.80\,\posdelta{6.30} & \cellcolor{oursbg}84.07\,\posdelta{6.57} \\
\midrule

\multirow{2}{*}{Spreadsheet}
& SkillOpt & 39.16 & 64.87\,\posdelta{25.71} & 53.21\,\posdelta{14.05} \\
& \cellcolor{oursbg}\method{} & \cellcolor{oursbg}39.16 & \cellcolor{oursbg}69.40\,\posdelta{30.24} & \cellcolor{oursbg}71.78\,\posdelta{32.62} \\
\midrule

\multirow{2}{*}{DocVQA}
& SkillOpt & 79.05 & 89.30\,\posdelta{10.25} & 89.30\,\posdelta{10.25} \\
& \cellcolor{oursbg}\method{} & \cellcolor{oursbg}79.05 & \cellcolor{oursbg}90.37\,\posdelta{11.32} & \cellcolor{oursbg}89.83\,\posdelta{10.78} \\

\bottomrule[1.5pt]
\end{tabularx}

\vspace{-5pt}
\end{table*}

\section{Discussion}
Building on the comparative results in Section~\ref{sec:experiments}, we further examine how graph structure supports skill execution and optimization, and whether the resulting skills transfer across models. We organize the discussion around three research questions (RQs):
\begin{itemize}[leftmargin=*]
    \item \textbf{RQ1--Graph Representation for Skill Execution} (Section~\ref{subsec:graph_representation}): Does explicit graph structure improve the execution of optimized skills?
    \item \textbf{RQ2--Graph Representation for Skill Optimization} (Section~\ref{subsec:ablation}): Does graph structure help discover higher-quality skills, and how do mutation and crossover contribute?
    \item \textbf{RQ3--Transferability of Graph-Structured Skills} (Section~\ref{subsec:transfer}): Do the skills optimized by \method{} remain effective when transferred to another LLM?
\end{itemize}

\subsection{Effect of Graph-Structured Representation on Skill Execution}
\label{subsec:graph_representation}

\textbf{Explicit Workflow Guidance Improves Skill Execution.}
To examine the role of graph structure during execution, we take the skills optimized by \method{} with GPT-5.4-nano and construct unstructured counterparts by removing explicit workflow organization while retaining global guidance and node-level instructions.
We evaluate both versions with GPT-5.4-nano on all five benchmarks.
As shown in Table~\ref{tab:graph}, removing graph structure reduces success rates by 4.52, 2.50, 4.19, 1.35, and 0.75 percentage points on SearchQA, Spreadsheet, DocVQA, LiveMath, and ALFWorld, respectively.
These consistent decreases suggest that global guidance and node-level instructions alone do not capture the full benefits of a graph-structured skill.
Explicitly organizing these instructions into context-specific workflows helps the agent apply them more effectively during execution.

\subsection{Advantage of Graph Representation in Skill Optimization}
\label{subsec:ablation}

To understand how graph structure facilitates skill optimization, we conduct ablation studies on GPT-5.4-nano and report the average results over three repeated experiments in Table~\ref{tab:ablation}. All ablation variants use the same experimental settings, differing only in the component ablated. When mutation or crossover is removed, we still generate \(N\) new skills per generation by cycling through the remaining operators.


\textbf{Graph structure benefits skill optimization.}
The \textit{w/o graph structure} variant initializes and evolves unstructured skills. Removing the graph structure decreases the average performance from 71.52 to 64.08, showing that population-based evolution alone is insufficient. The graph representation organizes skills into explicit procedural components and dependencies, providing a more structured search space for optimization.

\textbf{Crossover enables broader exploration.}
The \textit{w/o crossover} variant retains the graph representation, population, and mutation, but removes information exchange across candidates.
It can therefore be viewed as multiple parallel \emph{SkillOpt-style self-refinement} trajectories.
Its performance drops to 66.59, suggesting that crossover is important for combining effective components discovered along different search trajectories and enabling broader exploration beyond iterative
self-refinement.

\textbf{Mutation enables trajectory-driven refinement.}
The \textit{w/o mutation} variant removes the mutation operators and relies solely on crossover for skill evolution, resulting in the largest performance drop, to 54.50.
This shows that execution feedback is important for locally refining individual skills, while crossover complements this refinement through cross-candidate recombination.

\subsection{Transferring Skills Across LLMs}
\label{subsec:transfer}

\textbf{Optimized Skills Remain Effective Across LLMs.}
To evaluate cross-model transferability, we optimize skills with GPT-5.4-nano and deploy them on GPT-5.4.
For both \method{} and SkillOpt, Table~\ref{tab:transfer-performance} compares execution without a skill, with a skill optimized directly on GPT-5.4, and with a skill transferred from GPT-5.4-nano.
This comparison evaluates whether optimized procedural guidance remains useful beyond the model used for optimization.
Transferred skills outperform the no-skill baseline on all three benchmarks.
Moreover, transferred \method{} skills achieve higher scores than transferred SkillOpt skills, with the largest advantage on SpreadsheetBench.
On this benchmark, the transferred \method{} skill achieves 71.78, exceeding both its directly optimized counterpart (69.40) and the transferred SkillOpt skill (53.21).
These results demonstrate that skills optimized by \method{} can be reused across the evaluated models while retaining effective procedural guidance.

\section{Conclusion}

In this paper, we formulate agent skills as graph-structured natural-language artifacts that make workflow guidance explicit and organize reusable execution steps into a structured search space.
Building on this representation, we introduce \method{}, a population-based evolutionary computation framework with structure-aware mutation and crossover for refining and recombining procedural components.
Experiments across five agent benchmarks, two LLMs, and two execution settings demonstrate improved average performance over SkillOpt, while our token analysis shows lower aggregate optimization-token consumption.
Further analyses support the benefits of graph structure for both skill execution and evolutionary optimization, and demonstrate effective skill transfer from GPT-5.4-nano to GPT-5.4.

Future work includes combining our method with parametric optimization methods, extending the framework to richer graph composition mechanisms, and developing methods for merging graph-structured skills from diverse domains.

\bibliography{iclr2027_conference}
\bibliographystyle{iclr2027_conference}

\newpage
\appendix

\startcontents[sections]
\section*{Appendix Contents}
\printcontents[sections]{}{0}{}

\newpage

\section{Related Work}



\subsection{Agent Skill Optimization}

A skill encapsulates reusable procedural knowledge, including tool-use policies, applicability conditions, execution routines, and supporting resources \citep{li2026skillsbench,jiang2026sok}.
EvoSkill, Trace2Skill, SkillX, and AutoRefine use textual feedback obtained from agent execution trajectories to diagnose failures and improve skills \citep{alzubi2026evoskill,ni2026trace2skill,wang2026skillx,qiu2026autorefine}.
Meanwhile, SkillOpt studies how to train skills with deep-learning-style controls \citep{yang2026skillopt}.
In a different line, EvolveR, SAGE, and SKILLRL iteratively co-evolve the LLM and skills through reinforcement learning \citep{wu2025evolver,wang2026reinforcement,xia2026skillrl}.
AutoSkill constructs personalized skills from lifelong experience \citep{yang2026autoskill}, whereas SkillClaw constructs cross-user skills through aggregating interaction trajectories from multiple users \citep{ma2026skillclaw}.
Despite adopting different approaches to skill refinement, existing methods generally represent skills as unstructured natural-language instructions, which often lack workflow-level guidance, introduce substantial redundancy, and leave the optimizer with a large search space. In contrast, we formulate skills as graph-structured natural-language artifacts and optimize them with a population-based evolutionary framework. This design provides explicit workflow guidance, reduces redundancy, and makes skill optimization more tractable.

\subsection{Graph for Agent Skill}

Recent studies have begun to introduce graph for agent skills, mainly to improve skill retrieval and composition. Graph-of-Skills and SkillDAG incorporate graph structure into skill retrieval over large skill libraries, enabling agents to efficiently identify the subset of skills required to execute the current task \citep{liu2026graph,bai2026skilldag}. GraSP introduces graph-structured skill composition for large skill libraries, enabling agents to more effectively organize and orchestrate multiple skills during task execution \citep{xia2026grasp}. 
\citet{liang2026skill} convert existing skills to Scheduling-Structural-Logical (SSL) representations, facilitating skill retrieval from large skill libraries and skill risk assessment.
HiSkill discovers multiple skills and organizes them into a graph in which each skill corresponds to a node, and retrieves a task-relevant skill subset during task solving \citep{hao2026hiskill}.
These studies mainly focus on skill retrieval over large skill libraries and skill composition during task solving. 
In contrast, our work represents each skill as a graph-structured natural-language artifact and optimizes skills within this representation through structure-aware mutation and crossover to discover higher-quality skills.

\newpage
\section{Methodological Details}
\subsection{Benchmarks}
\label{app:benchmarks}


We provide the detailed introduction and settings of each benchmark in this subsection.

\paragraph{SearchQA.}
SearchQA \citep{dunn2017searchqa} evaluates question answering from accompanying textual evidence. Given a question and its evidence, the agent produces the answer to the question.

\paragraph{SpreadsheetBench.}
SpreadsheetBench \citep{ma2024spreadsheetbench} evaluates programmatic manipulation of real \texttt{.xlsx} workbooks. The agent generates Python code in an execution environment that provides the standard library, \texttt{openpyxl}, and \texttt{pandas}. We follow an iterative protocol in which the generated code is executed after each round and the resulting output or execution-level failure diagnostics is returned to the agent, which may revise the code in a subsequent round. We permit up to 30 code-generation rounds in the no-harness setting; with the Codex harness, each task uses a single code-generation round.

\paragraph{DocVQA.}
DocVQA \citep{mathew2021docvqa} is a visual question-answering task over document images. The agent receives a document image and a question, and returns the answer supported by the image.

\paragraph{LiveMathematicianBench (LiveMath).}
LiveMathematicianBench \citep{he2026livemathematicianbench} consists of mathematical multiple-choice problems. For each problem, the agent selects and outputs one of the provided answer options.

\paragraph{ALFWorld.}
ALFWorld \citep{shridhar2020alfworld} evaluates interaction in a persistent, text-based household environment. At each step, the agent observes the current state, selects an admissible action, and receives the next environment observation. Each episode is limited to 50 interaction steps.

\subsection{Dataset Splits}
\label{app:dataset_splits}

For each benchmark, we construct three disjoint sets. 
Following SkillOpt~\citep{yang2026skillopt}, all experiments use the same deterministic dataset partitioning procedure with \texttt{split\_seed=42}.
All baselines use exactly the same three dataset partitions to ensure a fair comparison.
The training set is used only for collecting rollout trajectories and failure feedback during optimization. 
The validation set is used to score candidate skills and guide population selection. The test set is reserved for final evaluation.
Table~\ref{tab:dataset-splits} reports the size of each set used in all experiments.

\begin{table}[H]
\centering
\small
\setlength{\tabcolsep}{30pt}
\renewcommand{\arraystretch}{1.08}
\caption{Sizes of the training, validation, and test sets used in the experiments. 
The training set is
used only for collecting rollout trajectories and failure feedback during
optimization.
The validation set is used
for skill selection during optimization, while the test set is used only for
final reporting.}
\label{tab:dataset-splits}
\begin{tabular}{l c c c}  
\toprule[1.5pt]
Benchmark & Train & Validation & Test \\
\midrule
SearchQA & 400 & 200 & 1400 \\
SpreadsheetBench & 80 & 40 & 280  \\
DocVQA & 107 & 53 & 374 \\
LiveMath & 35 & 18 & 124 \\
ALFWorld & 39 & 18 & 134 \\
\bottomrule[1.5pt]
\end{tabular}
\end{table}

\newpage

\subsection{Graph Structure Validation}
\label{app:graph_structure_validation}

To maintain the graph structure of skills during evolution, we apply a validator to every newly generated skill. The validator is a script that checks whether the skill follows the required graph schema and whether the declared nodes and the overall graph are structurally consistent with one another. 
For example, it verifies that every node referenced in a workflow
is defined in $V_s$ and that every declared node appears in at
least one workflow.
Generated skills that fail validation are discarded and regenerated. This validator helps keep the optimized skills well-formed and graph-structured throughout evolution.

\subsection{Complete Optimization Algorithm}
\label{app:optimization_algorithm}

Algorithm \ref{alg:complete-optimization} provides pseudocode for the proposed \method{} method.

\begin{breakablealgorithm}

\caption{Complete optimization algorithm of \method{}.}
\label{alg:complete-optimization}
\begin{algorithmic}[1]
\REQUIRE Initial graph skill $s_1$, train set $D_{\mathrm{train}}$,
validation set $D_{\mathrm{val}}$, population size $N$, generations $T$,
training sample size $B$, failure budget $K$
\ENSURE Optimized graph-structured skill $\hat{s}$
\STATE Initialize population $\mathcal{P}^{(1)} \leftarrow \{s_1\}$
\FOR{$i=1$ to $N-1$}
    \STATE Generate a candidate graph skill with the initialization prompt
    \WHILE{the candidate fails graph structure validation}
        \STATE Discard the candidate and regenerate it
    \ENDWHILE
    \STATE Add the candidate to $\mathcal{P}^{(1)}$
\ENDFOR
\FOR{each $s \in \mathcal{P}^{(1)}$}
    \STATE Evaluate $s$ on the full $D_{\mathrm{val}}$ and store fitness
    $J_{D_{val}}(s)$
\ENDFOR
\FOR{$t=1$ to $T$}
    \STATE Sample $B$ instances from $D_{\mathrm{train}}$
    \FOR{each parent $s \in \mathcal{P}^{(t)}$}
        \STATE Execute $s$ on the sampled instances
        \STATE Collect at most $K$ failed trajectories as reflection information
    \ENDFOR
    \STATE Sort $\mathcal{P}^{(t)}$ by validation fitness
    \STATE Assign parent-sampling weights $p_i \propto 1/(r_i+N)$, where
    $r_i$ is the validation rank
    \STATE $\widetilde{\mathcal{P}}^{(t+1)} \leftarrow \emptyset$
    \FOR{$j=1$ to $N$}
        \STATE Select the next operator in round-robin order from
        \{global mutation, graph mutation, global crossover, graph crossover\}
        \STATE Sample parent(s) according to the rank weights
        \STATE Invoke the selected operator prompt with parent skill(s) and,
        for mutation, the selected parent's reflection information
        \STATE Generate the new skill
        \WHILE{the new skill fails graph structure validation}
            \STATE Discard the candidate and regenerate it
        \ENDWHILE
        \STATE Add the new skill to $\widetilde{\mathcal{P}}^{(t+1)}$
    \ENDFOR
    \FOR{each child $s \in \widetilde{\mathcal{P}}^{(t+1)}$}
        \STATE Evaluate $s$ on the full $D_{\mathrm{val}}$ and store
        fitness $J_{D_{val}}(s)$
    \ENDFOR
    \STATE $\mathcal{P}^{(t+1)} \leftarrow$ top-$N$ skills from
    $\mathcal{P}^{(t)} \cup \widetilde{\mathcal{P}}^{(t+1)}$ by validation fitness
\ENDFOR
\STATE \textbf{return} $\hat{s}\in\arg\max_{s \in \mathcal{P}^{(T+1)}} J_{D_{val}}(s)$
\end{algorithmic}
\end{breakablealgorithm}

\newpage
\section{Additional Experiments}




\subsection{Significance Test}
\label{app:significance_test}

To examine whether there is a significant difference between \method{} and SkillOpt, we conduct a separate robustness experiment and use p-values from one-sided Welch's t-tests to assess whether \method{} significantly outperforms the promising skill optimization method SkillOpt.
For each benchmark, we report the test results of five individual skill optimization runs together with the mean, standard deviation, and p-value.
The procedural benchmarks show the strongest effect, with Spreadsheet and ALFWorld both achieving p-values below 0.05 and thus indicating GraphSkillEvo leads compared to SkillOpt. 
In contrast, the question-answering benchmarks show more modest gains, with SearchQA and DocVQA falling in the 0.05 to 0.10 range.

\begin{table}[H]
\centering
\footnotesize
\setlength{\tabcolsep}{0pt}
\renewcommand{\arraystretch}{1.12}
\caption{The significance test between SkillOpt and \method{} using GPT-5.4-nano without an agent harness. Avg and Std denote the mean and standard deviation, respectively. The reported p-values are computed using one-sided Welch's t-tests.}
\label{tab:significance-test}
\begin{tabularx}{\textwidth}{@{}%
>{\raggedright\arraybackslash}p{1.55cm}%
>{\raggedright\arraybackslash}p{2.35cm}%
*{7}{>{\centering\arraybackslash}X}%
>{\centering\arraybackslash}p{1.75cm}%
@{}}
\toprule[1.5pt]
Benchmark & Method & Run1 & Run2 & Run3 & Run4 & Run5 & Avg & Std & p-value \\
\midrule
\multirow{2}{*}{SearchQA}
& SkillOpt & 67.85 & 70.21 & 69.92 & 70.64 & 68.21 & 69.37 & 1.25 & \multicolumn{1}{c}{} \\
& \cellcolor{oursbg}\method{} & \cellcolor{oursbg}74.57 & \cellcolor{oursbg}69.35 & \cellcolor{oursbg}76.78 & \cellcolor{oursbg}73.07 & \cellcolor{oursbg}68.14 & \cellcolor{oursbg}72.38 & \cellcolor{oursbg}3.59 & \cellcolor{oursbg}0.068734203 \\
\midrule
\multirow{2}{*}{Spreadsheet}
& SkillOpt & 46.78 & 45.35 & 42.85 & 50.35 & 54.64 & 47.99 & 4.60 & \multicolumn{1}{c}{} \\
& \cellcolor{oursbg}\method{} & \cellcolor{oursbg}59.64 & \cellcolor{oursbg}59.28 & \cellcolor{oursbg}63.57 & \cellcolor{oursbg}59.64 & \cellcolor{oursbg}46.07 & \cellcolor{oursbg}57.64 & \cellcolor{oursbg}6.70 & \cellcolor{oursbg}0.016219296 \\
\midrule
\multirow{2}{*}{ALFWorld}
& SkillOpt & 41.79 & 47.76 & 57.46 & 54.47 & 58.95 & 52.09 & 7.18 & \multicolumn{1}{c}{} \\
& \cellcolor{oursbg}\method{} & \cellcolor{oursbg}67.16 & \cellcolor{oursbg}60.44 & \cellcolor{oursbg}58.20 & \cellcolor{oursbg}61.19 & \cellcolor{oursbg}53.73 & \cellcolor{oursbg}60.14 & \cellcolor{oursbg}4.88 & \cellcolor{oursbg}0.038218293 \\
\midrule
\multirow{2}{*}{DocVQA}
& SkillOpt & 75.40 & 77.54 & 75.66 & 73.52 & 76.20 & 75.66 & 1.46 & \multicolumn{1}{c}{} \\
& \cellcolor{oursbg}\method{} & \cellcolor{oursbg}78.87 & \cellcolor{oursbg}77.00 & \cellcolor{oursbg}82.35 & \cellcolor{oursbg}74.06 & \cellcolor{oursbg}77.54 & \cellcolor{oursbg}77.96 & \cellcolor{oursbg}3.02 & \cellcolor{oursbg}0.088788889 \\
\bottomrule[1.5pt]
\end{tabularx}
\end{table}

\subsection{Case Study}
\label{app:case_study}

\begin{figure}[t]
    \centering
    \includegraphics[width=0.95\linewidth]{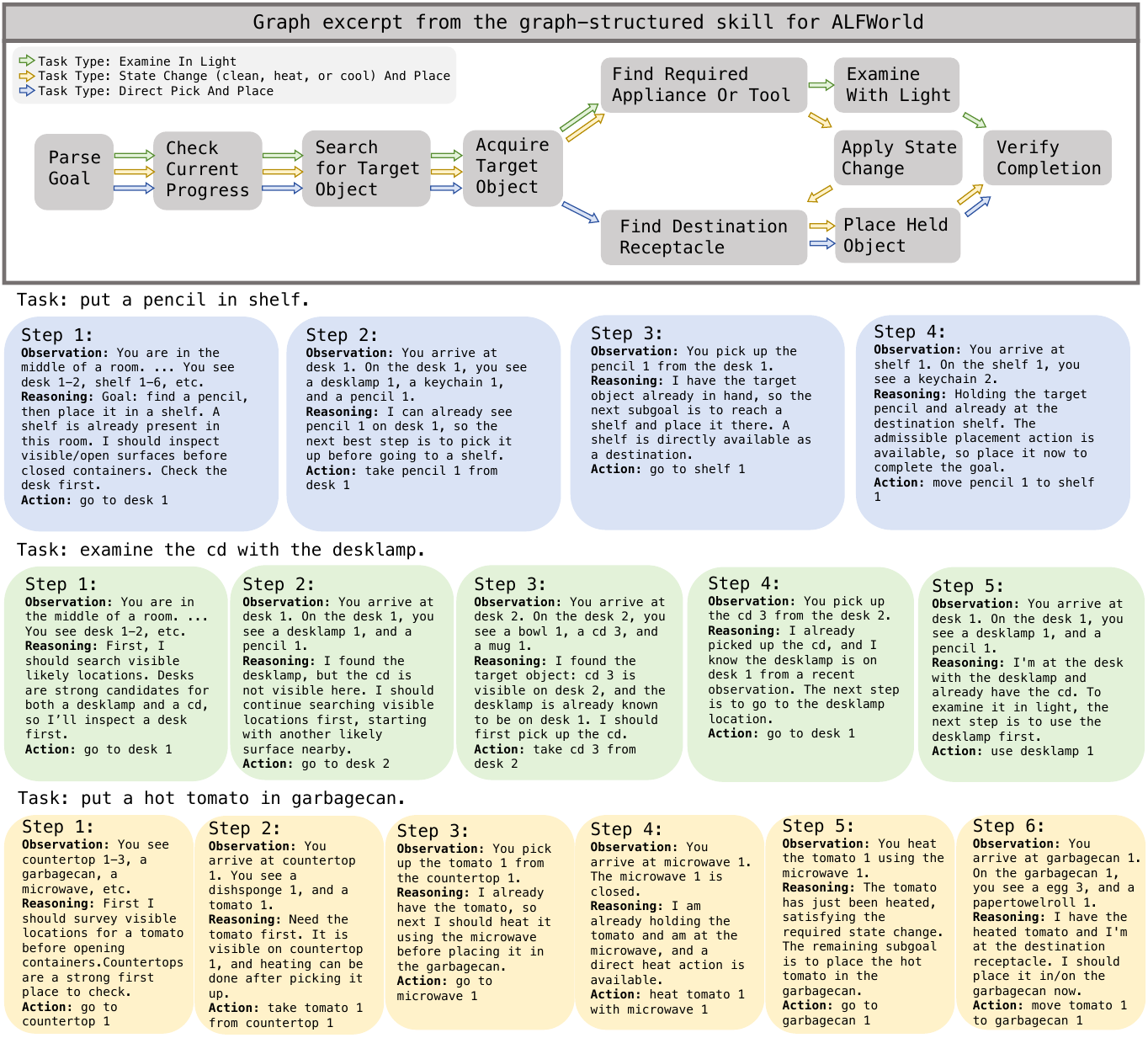}

    \caption{A case study of a graph-structured skill on ALFWorld. The figure presents a graph excerpt from the graph-structured skill for ALFWorld together with three representative execution trajectories. Different colors distinguish the three task types and indicate their corresponding workflow guidance and execution trajectories. The examples illustrate how graph-structured skills provide clear and explicit workflow guidance across different situations within the ALFWorld task.}
    \label{fig:case_study}
\end{figure}

To illustrate how graph-structured skills guide agent execution, we provide a case study on ALFWorld. Figure~\ref{fig:case_study} presents a graph excerpt from the graph-structured skill for ALFWorld together with three representative execution trajectories. 

\section{Baselines \& Licenses}

\subsection{Baseline Implementation Details}
\label{app:baseline_detail}

\textbf{No skill.}
The no-skill baseline evaluates the agent with an empty skill artifact.
No additional procedural guidance is prepended beyond the benchmark's native
task prompt.

\textbf{Human skill.}
The human-skill baseline uses a manually written benchmark-specific skill. The
skill is fixed during evaluation and is not optimized.

\textbf{LLM skill.}
The LLM-skill baseline uses a one-shot skill generated by GPT-5.4 from the
benchmark task description. It does not use evolutionary optimization or
validation feedback, and the generated skill is fixed during evaluation.

\textbf{SkillOpt.}
SkillOpt optimizes skills using rollout reflection, textual edit
selection, skill updating, and validation gating.
Across benchmarks, optimization runs
for 4 epochs. Each rollout batch contains 40 examples, with accumulation set to
1. The reflection minibatch size is 8, and the merge batch size is also 8.
For skill editing, the edit budget is 4 and the minimum edit budget is 2. The
edit budget follows a cosine schedule.
Slow update is enabled with 20 samples, and slow-update acceptance is also
controlled by validation gating. Meta-skill memory is enabled.

\subsection{Licenses}
\label{app:licenses}
The licenses and URLs of baselines are listed in Table \ref{tab:licenses}.

\begin{table}[H]
\centering
\scriptsize
\setlength{\tabcolsep}{10pt}
\renewcommand{\arraystretch}{1.2}
\caption{Links and licenses for datasets and method code.}
\label{tab:licenses}
\begin{tabularx}{\textwidth}{
    p{0.18\textwidth}
    >{\raggedright\arraybackslash}X
    >{\raggedright\arraybackslash}p{0.25\textwidth}
}
\toprule[1.5pt]
Resource & Link & License \\
\midrule
SearchQA & \url{https://huggingface.co/datasets/lucadiliello/searchqa} & Not specified \\
SpreadsheetBench & \url{https://huggingface.co/datasets/KAKA22/SpreadsheetBench} & CC-BY-SA-4.0 \\
DocVQA & \url{https://huggingface.co/datasets/lmms-lab/DocVQA} & Apache-2.0 on source card \\
LiveMathematicianBench & \url{https://huggingface.co/datasets/LiveMathematicianBench/LiveMathematicianBench} & Not specified \\
ALFWorld & \url{https://github.com/alfworld/alfworld} & MIT \\
SkillOpt & \url{https://github.com/microsoft/SkillOpt} & MIT \\
\bottomrule[1.5pt]
\end{tabularx}
\end{table}

\newpage

\section{Prompts and Optimized Skill Examples}
\label{app:prompts_skills}

This section presents the prompts used for skill initialization and evolutionary operators, together with an optimized skill example.
The prompts and skill example retain the section names used in
our implementation.
The \texttt{Global Guidance} and \texttt{Node Lists} sections specify $h_s$ and $V_s$, respectively.
The \texttt{Task Graphs} section specifies workflows through their applicability conditions and ordered node sequences.
Consecutive nodes in these workflows define the directed edges in $E_s$. The term \texttt{Task Graphs} names these workflows rather than an additional graph.

\subsection{Prompts for Evolutionary Operators}
\label{app:operator_prompts}

\method{} employs four evolutionary
operator prompts to optimize graph-structured skills. The four
operators are global-guidance mutation, graph-structure mutation,
global-guidance crossover, and graph-structure crossover. All
prompts require the LLM to return a complete graph-structured skill. The rest of this subsection provides the prompts of these four operators.

\begin{itemize}[leftmargin=*]
    \item \textbf{Global-guidance mutation.} This operator revises the global guidance of a selected parent skill according to its reflection information. 

    \begin{dialogbox}[Prompt for Global-guidance Mutation]
    \begin{lstlisting}[
      basicstyle=\scriptsize\ttfamily,
      breaklines=true,
      columns=fullflexible,
      keepspaces=true,
      showstringspaces=false,
      numbers=none,
      xleftmargin=0pt
    ]
You are a reflection-driven non-graph revision operator for graph-structured skill documents.

A graph-structured skill document contains:
- a `## Global Guidance` section, which provides general guidance, principles, and output format, and graph-use instructions;
- a `## Node Lists` section, where each node represents a reusable subtask, reasoning step, tool-use step, validation step, or recovery strategy;
- a `## Task Graphs` section, where each task graph describes workflow paths that connect node names into executable task-solving procedures.

Your job is to revise only the `## Global Guidance` section.

Review all provided evaluation results and failed-trajectory reflections to identify the prevalent recurring failure patterns and any missing, misleading, ignored, or redundant global guidance. Revise the global guidance to address the observed gaps while avoiding duplication in existing guidance.

You may:
- refine the Overview;
- improve output-format instructions;
- add or revise General Principles;
- clarify how the skill should reason, verify, and recover from mistakes;
- update the graph-structured skill execution guide.

Do not intentionally change the `## Node Lists` or `## Task Graphs` sections. The implementation will enforce this boundary, but your response should respect it.

Keep section boundaries and structure:
- `## Global Guidance` should remain a single section, organized with concise markdown subsections when useful;
- it should contain guidance that applies across all nodes and workflows, not reminders or requirements for one specific node;
- it usually contains **`### General Principles`**, which lists concise, portable rules that should guide all workflows;
- it usually contains **`### Graph-structured Skill Execution Guide`**, which tells the agent how to use the graph-structured skill: select a task graph, follow its exact node names in order, and apply the matching node instructions;
- do not add hidden workflows, task graph paths, node definitions, or node-specific procedural details to `## Global Guidance`;

Remove redundancy:
- merge duplicate or overlapping guidance.
Do not include file paths, IDs, gold values, entity names, or dataset-specific memorized facts.

Before returning, verify that `## Global Guidance` is concise, non-redundant, and contains only global guidance.
Avoid mentioning that the result was produced by mutation.
The returned `new_skill` must still be a complete graph-structured skill document containing:
- `## Global Guidance`,
- `## Node Lists`,
- `## Task Graphs`.

Return JSON only:
{"new_skill": "full skill document", "notes": ["brief change note"]}
    \end{lstlisting}
    \end{dialogbox}

    \item \textbf{Graph-structure mutation.} This operator revises the reusable
    nodes and task workflows of a selected parent skill using its reflection
    information. It may refine node instructions, add or remove nodes, and
    adjust task workflows while preserving the parent's global guidance and
    maintaining the graph structure.

    \begin{dialogbox}[Prompt for Graph-structure Mutation]
    \begin{lstlisting}[
      basicstyle=\scriptsize\ttfamily,
      breaklines=true,
      columns=fullflexible,
      keepspaces=true,
      showstringspaces=false,
      numbers=none,
      xleftmargin=0pt
    ]
You are a reflection-driven graph-structure revision operator for graph-structured skill documents.

A graph-structured skill document contains:
- a `## Global Guidance` section, which provides general guidance, principles, and output format;
- a `## Node Lists` section, where each node represents a reusable subtask, reasoning step, tool-use step, validation step, or recovery strategy;
- a `## Task Graphs` section, where each task graph describes workflow paths that connect node names into executable task-solving procedures.

Your job is to revise only the graph structure of the selected parent skill: the `## Node Lists` section and the `## Task Graphs` section.

Review all provided evaluation results and failed-trajectory reflections to identify the prevalent recurring failure patterns and any missing, misleading, ignored, or redundant graph steps. Revise the graph to address the observed gaps while avoiding duplication in existing nodes or workflows.

You may:
- refine node instructions to make them more concrete and actionable;
- add reusable nodes for missing subtasks, checks, fallback behavior, or failure-handling steps;
- merge redundant nodes;
- remove obsolete, misleading, or unused nodes;
- reorder or reroute task graph paths;
- add, delete, split, merge, or adjust workflow branches.

Keep the `## Global Guidance` section unchanged. 

Keep section boundaries and structure:
- `## Node Lists` should be structured as one `### <Node Name>` heading per reusable execution step, followed by the instructions the agent should follow for that node;
- `## Task Graphs` should be structured as task-type workflows, each with a `**Use when:**` condition and a `**Workflow:**` list;
- workflow items must be exact node names only;

Maintain graph consistency:
- every node referenced in `## Task Graphs` must appear as a node heading in `## Node Lists`;
- every node in `## Node Lists` must be used by at least one task graph workflow;
- deleted or merged nodes must not remain in any task graph path;
- node names must be exactly consistent between node headings and graph paths;
- task graph paths should remain executable, ordered, and non-contradictory.

Remove redundancy:
- merge duplicate or near-duplicate nodes before returning;
- merge overlapping workflows;
- remove repeated or equivalent instructions inside nodes, across nodes, or across workflows;

Do not include file paths, IDs, gold values, entity names, or dataset-specific memorized facts.

Avoid mentioning that the result was produced by mutation.

The returned `new_skill` must still be a complete graph-structured skill document containing:
- `## Global Guidance`,
- `## Node Lists`,
- `## Task Graphs`.

Return JSON only:
{"new_skill": "full skill document", "notes": ["brief change note"]}
    \end{lstlisting}
    \end{dialogbox}

    \item \textbf{Global-guidance crossover.} This operator recombines useful
    global guidance from two selected parent skills while preserving the graph structure of one parent.

    \begin{dialogbox}[Prompt for Global-guidance Crossover]
    \begin{lstlisting}[
      basicstyle=\scriptsize\ttfamily,
      breaklines=true,
      columns=fullflexible,
      keepspaces=true,
      showstringspaces=false,
      numbers=none,
      xleftmargin=0pt
    ]
You are a non-graph crossover operator for graph-structured skill documents.

A graph-structured skill document contains:
- a `## Global Guidance` section, which provides general guidance, principles, and output format;
- a `## Node Lists` section, where each node represents a reusable subtask, reasoning step, tool-use step, validation step, or recovery strategy;
- a `## Task Graphs` section, where each task graph describes workflow paths that connect node names into executable task-solving procedures.

Your job is to perform crossover only on the `## Global Guidance` section.

Compare Parent A and Parent B and identify useful non-graph guidance.

Then recombine the non-graph guidance into one complete child skill. You may:
- import useful principles from Parent B;
- replace weak or vague guidance from Parent A;
- merge complementary instructions;
- remove duplicated, conflicting, or overly specific prose;
- improve clarity and concision.

Do not intentionally change Parent A's `## Node Lists` or `## Task Graphs` sections. The implementation will enforce this boundary, but your response should respect it.

Keep section boundaries and structure:
- `## Global Guidance` should remain a single section, organized with concise markdown subsections when useful;
- it should contain guidance that applies across all nodes and workflows, not reminders or requirements for one specific node;
- it usually contains **`### General Principles`**, which lists concise, portable rules that should guide all workflows;
- it usually contains **`### Graph-structured Skill Execution Guide`**, which tells the agent how to use the graph-structured skill: select a task graph, follow its exact node names in order, and apply the matching node instructions;
- do not add hidden workflows, task graph paths, node definitions, or node-specific procedural details to `## Global Guidance`;

Remove redundancy:
- merge duplicate or overlapping guidance.

Preserve concrete, actionable guidance. Do not include file paths, IDs, gold values, entity names, or dataset-specific memorized facts.

Before returning, verify that `## Global Guidance` is concise, non-redundant, and contains only global guidance.

Avoid mentioning that the result was produced by crossover.

The returned `new_skill` must still be a complete graph-structured skill document containing:
- `## Global Guidance`,
- `## Node Lists`,
- `## Task Graphs`.

Return JSON only:
{"new_skill": "full skill document", "notes": ["brief change note"]}
    \end{lstlisting}
    \end{dialogbox}

    \item \textbf{Graph-structure crossover.} This operator recombines useful
    reusable nodes and task workflows from two selected parent skills while preserving the global guidance of one parent.

    \begin{dialogbox}[Prompt for Graph-structure Crossover]
    \begin{lstlisting}[
      basicstyle=\scriptsize\ttfamily,
      breaklines=true,
      columns=fullflexible,
      keepspaces=true,
      showstringspaces=false,
      numbers=none,
      xleftmargin=0pt
    ]
You are a graph-structure crossover operator for graph-structured skill documents.

A graph-structured skill document contains:
- a `## Global Guidance` section, which provides general guidance, principles, and output format;
- a `## Node Lists` section, where each node represents a reusable subtask, reasoning step, tool-use step, validation step;
- a `## Task Graphs` section, where each task graph describes workflow paths that connect node names into executable task-solving procedures.

Your job is to perform crossover only on the graph structure: the `## Node Lists` section and the `## Task Graphs` section.

Recombine the graph structure into one complete child skill. You may:
- import useful nodes from Parent B into Parent A's graph;
- replace weak nodes with stronger alternatives;
- merge overlapping nodes from both parents;
- remove redundant or conflicting nodes;
- exchange or recombine task graph paths;
- build a better workflow by combining complementary subgraph fragments.

Keep Parent A's `## Global Guidance` section unchanged. The implementation will enforce this boundary, but your response should respect it.

Keep section boundaries and structure:
- `## Node Lists` should be structured as one `### <Node Name>` heading per reusable execution step, followed by the instructions the agent should follow for that node;
- `## Task Graphs` should be structured as task-type workflows, each with a `**Use when:**` condition and a `**Workflow:**` list;
- workflow items must be exact node names only;

Maintain graph consistency:
- every node referenced in `## Task Graphs` must appear as a node heading in `## Node Lists`;
- every node in `## Node Lists` must be used by at least one task graph workflow;
- deleted or merged nodes must not remain in any task graph path;
- node names must be exactly consistent between node headings and graph paths;
- task graph paths should remain executable, ordered, and non-contradictory.

Remove redundancy:
- merge duplicate or near-duplicate nodes before returning;
- merge overlapping workflows;
- remove repeated or equivalent instructions inside nodes, across nodes, or across workflows;

Preserve concrete, actionable guidance. Do not include file paths, IDs, gold values, entity names, or dataset-specific memorized facts.

Avoid mentioning that the result was produced by crossover.

The returned `new_skill` must still be a complete graph-structured skill document containing:
- `## Global Guidance`,
- `## Node Lists`,
- `## Task Graphs`.

Return JSON only:
{"new_skill": "full skill document", "notes": ["brief change note"]}
    \end{lstlisting}
    \end{dialogbox}
\end{itemize}

\subsection{Prompt for Skill Initialization}
\begin{dialogbox}[Prompt for Skill Initialization]
    \begin{lstlisting}[
      basicstyle=\scriptsize\ttfamily,
      breaklines=true,
      columns=fullflexible,
      keepspaces=true,
      showstringspaces=false,
      numbers=none,
      xleftmargin=0pt
    ]
You create initial graph-structured skill documents for an agent benchmark.

Generate complete, diverse skill documents that can be used directly by the target agent.
Keep each skill self-contained and practical. Use the benchmark context to infer the main task types, design suitable workflows for them, and convert recurring execution steps into reusable graph nodes.

Every skill must preserve this graph-structured organization:
1. A `## Global Guidance` section containing instructions that apply across all nodes and workflows.
2. A `## Node Lists` section containing reusable subtask nodes. Node headings should be markdown headings such as `### Parse Goal`.
3. A `## Task Graphs` section containing task-type workflows that connect node names into executable task-solving procedures.

Use this markdown layout for each skill document:

`# <Skill Name>`

`## Global Guidance`

`## Node Lists`

`## Task Graphs`

The optional title may appear before `## Global Guidance`, but the three required sections must appear exactly in this order.

Section requirements:

### `## Global Guidance`
This section contains instructions that apply across all nodes and workflows, including how to use the graph-structured skill.

Organize it with concise markdown subsections. Prefer the following subsections when they are useful:
- `### Overview`: briefly state the benchmark and the agent's role.
- `### General Principles`: list concise global rules that should guide all workflows.
- `### Graph-structured Skill Execution Guide`: explain how to execute the skill as a graph: select the relevant task graph using its `Use when` condition, execute the listed node names in order, and apply the instructions under each matching node in `## Node Lists`.

The `### General Principles` subsection should contain portable rules, not a hidden workflow. Put step-by-step procedures, search actions, computations, validations, and recovery routines into `## Node Lists` nodes, then connect them in `## Task Graphs`.

The graph-execution guide should be short and explicit. It may state that:
- `## Global Guidance` applies across all nodes and workflows;
- `## Node Lists` defines reusable execution steps;
- `## Task Graphs` chooses and orders those steps for each task type;
- a numbered workflow should use `1. A`, `2. B`, `3. C` formatting, where `A`, `B`, and `C` are exact node names in execution order.

### `## Node Lists`
This section defines the node library. Each node is a reusable step that can appear in one or more workflows inside the task graphs.

Organize it as:
- one markdown heading per node, for example `### Parse Request`, `### Gather Evidence`, or `### Verify Answer`;
- under each node heading, concise bullets with instructions the agent should follow while executing that node;

Effective nodes should:
- have short, action-oriented names that can be referenced exactly from task graphs;
- represent reusable subtasks, reasoning steps, tool-use steps, validation steps, or recovery strategies;
- be specific enough to guide behavior, but general enough to transfer across benchmark instances;
- separate distinct responsibilities when the order matters, such as parsing the goal, locating evidence, computing, and validating;

Avoid nodes that are empty, redundant, purely decorative, or tied to a single example. Do not create a long block of prose under one node when several reusable nodes would make the graph clearer.

### `## Task Graphs`
This section defines task graphs as collections of workflows, where each workflow is an executable path made from node names in `## Node Lists`.

Organize it as one or more task-type subsections:
- each task graph should have a heading such as `### Direct Evidence Question` or `### Multi-Step Calculation`;
- include a `**Use when:**` line describing when that workflow applies;
- include a `**Workflow:**` block formatted as `1. A`, `2. B`, `3. C`, where `A`, `B`, and `C` are exact node names from `## Node Lists`;
- each numbered item should be an exact node name from `## Node Lists`;

Maintain graph consistency:
- every node referenced in `## Task Graphs` must appear as a node heading in `## Node Lists`;
- node names must match exactly between node headings and graph paths;
- every retained node should be useful for at least one task graph or clearly reusable;
- task graph paths should be ordered, executable, and non-contradictory.

Return JSON only:
{"skills": ["full skill document", "..."]}
    \end{lstlisting}
\end{dialogbox}

\subsection{Example of Optimized Graph-Structured Skills}
\label{app:graph-structured_skills}

This subsection presents an example of optimized graph-structured skills produced by \method.

\begin{dialogbox2}[Example of optimized graph-structured skill: Skill for Spreadsheet]
\begin{lstlisting}[
      basicstyle=\scriptsize\ttfamily,
      breaklines=true,
      columns=fullflexible,
      keepspaces=true,
      showstringspaces=false,
      numbers=none,
      xleftmargin=0pt
    ]
# Spreadsheet Formula, Lookup, and Reporting Skill

## Global Guidance

### Overview
Use this skill for SpreadsheetBench tasks that require formulas, lookups, cross-sheet transfers, totals, summaries, or light presentation changes tied to computed spreadsheet outputs. In this benchmark, the deliverable is the modified workbook, so prioritize making the target cells contain the correct final workbook results.

### Output Requirements
- Return only a single complete ```python ... ``` fenced block containing the full script.
- Do not include explanation, comments outside the code block, or partial / placeholder code.
- Read from `INPUT_PATH`, write the modified workbook to `OUTPUT_PATH`, and preserve unrelated workbook content.

### General Principles
- Use only `openpyxl` and `pandas`.
- Prefer `openpyxl` for preserving workbook structure, writing cells, and applying basic formatting.
- Do not rely on `openpyxl` to evaluate Excel formulas.
- In SpreadsheetBench cell-level tasks, prefer writing final static values whenever formula evaluation would otherwise be required to make the workbook show the correct result.
- If the user names an Excel method or function such as INDEX/MATCH, SUMIF, IF, FILTER, or lookup formulas, treat that as the logic to reproduce; you may compute the result in Python instead of writing formula strings unless the task clearly requires formulas to remain in the sheet.
- If final displayed values are what matter, writing unevaluated formulas is insufficient.
- Infer target ranges, anchors, and fill extents from the actual workbook, not only the truncated preview.
- Do not stop range filling at the first blank cell if surrounding structure shows the table or requested output continues.
- For matrix-style transfers, resolve both row keys and column keys from headers and fill the full applicable destination area.
- For list outputs, write all qualifying items contiguously in the destination region, including rows beyond currently populated output cells when needed.
- Normalize header and key matching conservatively: trim spaces, compare case-insensitively, and handle common text variations, dates, and symbols such as tick marks.
- Preserve unrelated formulas, formatting, and sheets.
- Apply only the formatting explicitly requested or clearly necessary for the requested output.

### Verification and Recovery
- Before saving, verify that the intended target cells were actually written.
- Reopen the output workbook and confirm the target cells contain concrete expected-type results: numbers, text, or blanks as requested, not just formula strings when static results are needed.
- Verify fill extent carefully, including the last applicable row or column.
- If a lookup or summary output seems empty, re-check sheet selection, header detection, destination anchors, and whether blank cells should be written as `0`, empty string, or left empty according to the instruction.

### Graph-structured Skill Execution Guide
- `## Global Guidance` applies to every node and workflow.
- `## Node Lists` defines reusable steps for formula and reporting tasks.
- `## Task Graphs` chooses the right ordered workflow for each task type.
- Select the graph using its `Use when` line, then execute each listed node in order.
- Follow node instructions exactly by node name.
- When a workflow offers a choice between writing formulas and computing results, choose the path that will make the saved workbook contain the correct final target-cell contents under SpreadsheetBench evaluation.

## Node Lists

### Parse Task And Output Requirements
- Classify the request as an existing-cell fill, cross-sheet lookup or join, summary/report generation, split-or-expand transformation, or direct static transformation into a destination area.
- Extract explicit target cells or ranges, source sheets, destination sheets, lookup keys, requested calculations, blank-on-error behavior, and any formatting requests.
- If the instruction is phrased as an Excel-formula question, treat the named formula as the logic to reproduce; do not assume the sheet must retain a live formula unless the request clearly requires that.

### Inspect Workbook Context
- Load the workbook and inspect the actual used range of each relevant sheet.
- Examine headers, neighboring columns, existing formulas, placeholder cells, lookup tables, summary areas, date/report cells, and nearby styles that may need to be matched.
- Confirm the real workbook structure directly rather than relying only on the preview.

### Identify Explicit Target Cells
- Resolve the exact answer cell, output range, or destination block named by the instruction, workbook labels, or benchmark answer position.
- For single-cell or short fixed-range tasks, center the script around writing those exact coordinates rather than only constructing general helpers.
- Record these target coordinates for later save-time and reopen-time verification.

### Resolve Headers Keys And Target Area
- Map user-described headers, labels, and columns to actual workbook columns using conservative normalized matching.
- Resolve exact source and destination sheets, key columns, value columns, date columns, selection cells, and the area that should be overwritten.
- For cross-sheet work, confirm how source rows align to destination rows and identify any interval fields such as start and end dates.

### Determine Fill Extent And Rewrite Strategy
- Infer the full row or column extent from related populated columns, existing table structure, templates, and the requested output area.
- Decide whether the task is an in-place fill, a generated report block, or a rewritten exploded output that should replace prior contents.
- When regenerating a destination block, plan to clear stale formulas or old rows that would otherwise remain below or beside the new results.

### Choose Output Method
- Prefer static Python-computed values whenever unevaluated formulas would leave the saved workbook incorrect, especially for cell-level tasks.
- Use Excel formulas only when the instruction clearly requires formulas to remain in the sheet; even then, prepare equivalent Python logic for recovery.
- For direct-reference tasks or blank-if-source-blank behavior, simple value transfer may be the correct final method.

### Build Script Skeleton
- Build a complete executable script using `openpyxl` and `pandas`, reading from `INPUT_PATH` and writing to `OUTPUT_PATH`.
- Organize the script into workbook loading, helper functions, source discovery, computation or formula writing, formatting, save, and verification steps.
- Ensure the returned script is complete and not truncated.

### Interpret Formula Logic In Python
- Translate the user-described spreadsheet logic into Python-ready rules before writing output values.
- Handle common patterns such as direct cell reference with blank propagation, OR-style multi-selection criteria, `All` meaning no filter, workday/date-window aggregation, lookup matching, and conditional blanks.
- Build reusable helpers for normalized text comparison, date coercion, aggregation, and default-value handling.

### Write Formula Pattern
- Only when formulas must remain, write the target formula or formula pattern with correct relative or absolute references and requested blank/error handling.
- If the pattern extends across a range, fill it across the full resolved extent.
- Use this node only after the equivalent Python fallback logic is understood.

### Compute And Write Static Results
- Compute final values directly in Python for exact lookups, multi-criteria matches, date-range checks, totals, derived metrics, category mapping, sorting, grouped outputs, and other requested transformations.
- Write concrete results into every target cell in the resolved extent, including exact benchmark answer cells for single-cell tasks.
- When no match is found or the instruction requests blanks, write blank, `0`, empty string, or leave empty exactly as required.

### Expand Records Into Output Rows
- Parse source fields that encode multiple values, conditions, or list members and expand them into the required row-wise output structure.
- Produce one destination row per required emitted result, copying base fields and leaving unspecified extracted fields blank.
- Preserve source-driven ordering or destination-template ordering when possible.

### Create Or Update Summary Output
- Create or locate the destination report or output block and write headers, grouped sections, or paired/list outputs only when the task calls for them.
- Use this node for summary sheets, compact report areas, and designated destination blocks for generated result lists.
- Preserve unrelated cells outside the requested output area.

### Clear Target Output Area
- Before rewriting generated lists, reports, or expanded outputs, clear obsolete contents in the destination block that would otherwise leave stale rows, leftover formulas, or partial prior results.
- Keep headers and unrelated regions intact.
- Use the resolved output area and prior populated extent to determine what should be cleared.

### Apply Requested Formatting
- Apply only the formatting explicitly requested, such as bold text, alignment, number formats, widths, freeze panes, or copying fill/style from adjacent reference cells.
- When formatting should match a nearby column or header, clone the needed style attributes from the reference cells rather than inventing new formatting.
- Preserve existing formatting elsewhere.

### Save Workbook
- Save the modified workbook to `OUTPUT_PATH`.

### Reopen Verify And Recover
- Reopen the saved workbook and inspect the explicit target cells, target range, and final fill extent directly.
- Confirm the workbook contains concrete expected-type outputs and that required cells are not left as `None` or unevaluated formulas when static results are needed.
- If verification fails, recompute from the Python logic, overwrite the target area with static values, clear stale leftover cells if necessary, resave, and recheck before finalizing.

## Task Graphs

### Cell-Level Static Fill Or Direct Reference
**Use when:** The task asks to fill one cell or an existing small range in place, including formula-like requests where the benchmark needs final displayed values rather than a live formula.

**Workflow:**
1. Parse Task And Output Requirements
2. Inspect Workbook Context
3. Identify Explicit Target Cells
4. Resolve Headers Keys And Target Area
5. Determine Fill Extent And Rewrite Strategy
6. Choose Output Method
7. Build Script Skeleton
8. Interpret Formula Logic In Python
9. Compute And Write Static Results
10. Apply Requested Formatting
11. Save Workbook
12. Reopen Verify And Recover

### Formula Must Remain In Sheet
**Use when:** The task clearly requires formulas to remain visible in the target cells or to preserve a formula pattern in the workbook.

**Workflow:**
1. Parse Task And Output Requirements
2. Inspect Workbook Context
3. Identify Explicit Target Cells
4. Resolve Headers Keys And Target Area
5. Determine Fill Extent And Rewrite Strategy
6. Choose Output Method
7. Build Script Skeleton
8. Interpret Formula Logic In Python
9. Write Formula Pattern
10. Apply Requested Formatting
11. Save Workbook
12. Reopen Verify And Recover

### Cross-Sheet Lookup Or Multi-Criteria Match
**Use when:** The task asks to pull, match, or aggregate values from another sheet or table using one or more keys, categories, flags, or date conditions.

**Workflow:**
1. Parse Task And Output Requirements
2. Inspect Workbook Context
3. Identify Explicit Target Cells
4. Resolve Headers Keys And Target Area
5. Determine Fill Extent And Rewrite Strategy
6. Choose Output Method
7. Build Script Skeleton
8. Interpret Formula Logic In Python
9. Compute And Write Static Results
10. Apply Requested Formatting
11. Save Workbook
12. Reopen Verify And Recover

### Summary Sheet Or Report Block
**Use when:** The task asks to create or update a summary sheet, grouped totals section, compact report area, or another dedicated generated output block.

**Workflow:**
1. Parse Task And Output Requirements
2. Inspect Workbook Context
3. Identify Explicit Target Cells
4. Resolve Headers Keys And Target Area
5. Determine Fill Extent And Rewrite Strategy
6. Choose Output Method
7. Build Script Skeleton
8. Create Or Update Summary Output
9. Clear Target Output Area
10. Interpret Formula Logic In Python
11. Compute And Write Static Results
12. Apply Requested Formatting
13. Save Workbook
14. Reopen Verify And Recover

### Split Or Expand Encoded Fields To Rows
**Use when:** The task asks to parse a source field and expand encoded values, repeated members, or condition fragments into multiple destination rows or columns in an output block.

**Workflow:**
1. Parse Task And Output Requirements
2. Inspect Workbook Context
3. Identify Explicit Target Cells
4. Resolve Headers Keys And Target Area
5. Determine Fill Extent And Rewrite Strategy
6. Choose Output Method
7. Build Script Skeleton
8. Expand Records Into Output Rows
9. Clear Target Output Area
10. Compute And Write Static Results
11. Apply Requested Formatting
12. Save Workbook
13. Reopen Verify And Recover
\end{lstlisting}
\end{dialogbox2}

\subsection{Example of an Unstructured Counterpart}
\label{app:unstructured_counterpart}

This subsection presents the unstructured counterpart of the graph-structured skill shown above, illustrating the conversion procedure described in Section \ref{subsec:graph_representation}. It is obtained by removing the explicit workflow organization while retaining the global guidance and node-level instructions.

\begin{dialogbox2}[Example of an unstructured counterpart of a graph-structured skill]
\begin{lstlisting}[
      basicstyle=\scriptsize\ttfamily,
      breaklines=true,
      columns=fullflexible,
      keepspaces=true,
      showstringspaces=false,
      numbers=none,
      xleftmargin=0pt
    ]
# Spreadsheet Formula, Lookup, and Reporting Skill

## Global Guidance

### Overview
Use this skill for SpreadsheetBench tasks that require formulas, lookups, cross-sheet transfers, totals, summaries, or light presentation changes tied to computed spreadsheet outputs. In this benchmark, the deliverable is the modified workbook, so prioritize making the target cells contain the correct final workbook results.

### Output Requirements
- Return only a single complete ```python ... ``` fenced block containing the full script.
- Do not include explanation, comments outside the code block, or partial / placeholder code.
- Read from `INPUT_PATH`, write the modified workbook to `OUTPUT_PATH`, and preserve unrelated workbook content.

### General Principles
- Use only `openpyxl` and `pandas`.
- Prefer `openpyxl` for preserving workbook structure, writing cells, and applying basic formatting.
- Do not rely on `openpyxl` to evaluate Excel formulas.
- In SpreadsheetBench cell-level tasks, prefer writing final static values whenever formula evaluation would otherwise be required to make the workbook show the correct result.
- If the user names an Excel method or function such as INDEX/MATCH, SUMIF, IF, FILTER, or lookup formulas, treat that as the logic to reproduce; you may compute the result in Python instead of writing formula strings unless the task clearly requires formulas to remain in the sheet.
- If final displayed values are what matter, writing unevaluated formulas is insufficient.
- Infer target ranges, anchors, and fill extents from the actual workbook, not only the truncated preview.
- Do not stop range filling at the first blank cell if surrounding structure shows the table or requested output continues.
- For matrix-style transfers, resolve both row keys and column keys from headers and fill the full applicable destination area.
- For list outputs, write all qualifying items contiguously in the destination region, including rows beyond currently populated output cells when needed.
- Normalize header and key matching conservatively: trim spaces, compare case-insensitively, and handle common text variations, dates, and symbols such as tick marks.
- Preserve unrelated formulas, formatting, and sheets.
- Apply only the formatting explicitly requested or clearly necessary for the requested output.

### Verification and Recovery
- Before saving, verify that the intended target cells were actually written.
- Reopen the output workbook and confirm the target cells contain concrete expected-type results: numbers, text, or blanks as requested, not just formula strings when static results are needed.
- Verify fill extent carefully, including the last applicable row or column.
- If a lookup or summary output seems empty, re-check sheet selection, header detection, destination anchors, and whether blank cells should be written as `0`, empty string, or left empty according to the instruction.

- Classify the request as an existing-cell fill, cross-sheet lookup or join, summary/report generation, split-or-expand transformation, or direct static transformation into a destination area.
- Extract explicit target cells or ranges, source sheets, destination sheets, lookup keys, requested calculations, blank-on-error behavior, and any formatting requests.
- If the instruction is phrased as an Excel-formula question, treat the named formula as the logic to reproduce; do not assume the sheet must retain a live formula unless the request clearly requires that.
- Load the workbook and inspect the actual used range of each relevant sheet.
- Examine headers, neighboring columns, existing formulas, placeholder cells, lookup tables, summary areas, date/report cells, and nearby styles that may need to be matched.
- Confirm the real workbook structure directly rather than relying only on the preview.
- Resolve the exact answer cell, output range, or destination block named by the instruction, workbook labels, or benchmark answer position.
- For single-cell or short fixed-range tasks, center the script around writing those exact coordinates rather than only constructing general helpers.
- Record these target coordinates for later save-time and reopen-time verification.
- Map user-described headers, labels, and columns to actual workbook columns using conservative normalized matching.
- Resolve exact source and destination sheets, key columns, value columns, date columns, selection cells, and the area that should be overwritten.
- For cross-sheet work, confirm how source rows align to destination rows and identify any interval fields such as start and end dates.
- Infer the full row or column extent from related populated columns, existing table structure, templates, and the requested output area.
- Decide whether the task is an in-place fill, a generated report block, or a rewritten exploded output that should replace prior contents.
- When regenerating a destination block, plan to clear stale formulas or old rows that would otherwise remain below or beside the new results.
- Prefer static Python-computed values whenever unevaluated formulas would leave the saved workbook incorrect, especially for cell-level tasks.
- Use Excel formulas only when the instruction clearly requires formulas to remain in the sheet; even then, prepare equivalent Python logic for recovery.
- For direct-reference tasks or blank-if-source-blank behavior, simple value transfer may be the correct final method.
- Build a complete executable script using `openpyxl` and `pandas`, reading from `INPUT_PATH` and writing to `OUTPUT_PATH`.
- Organize the script into workbook loading, helper functions, source discovery, computation or formula writing, formatting, save, and verification steps.
- Ensure the returned script is complete and not truncated.
- Translate the user-described spreadsheet logic into Python-ready rules before writing output values.
- Handle common patterns such as direct cell reference with blank propagation, OR-style multi-selection criteria, `All` meaning no filter, workday/date-window aggregation, lookup matching, and conditional blanks.
- Build reusable helpers for normalized text comparison, date coercion, aggregation, and default-value handling.
- Only when formulas must remain, write the target formula or formula pattern with correct relative or absolute references and requested blank/error handling.
- If the pattern extends across a range, fill it across the full resolved extent.
- Use this node only after the equivalent Python fallback logic is understood.
- Compute final values directly in Python for exact lookups, multi-criteria matches, date-range checks, totals, derived metrics, category mapping, sorting, grouped outputs, and other requested transformations.
- Write concrete results into every target cell in the resolved extent, including exact benchmark answer cells for single-cell tasks.
- When no match is found or the instruction requests blanks, write blank, `0`, empty string, or leave empty exactly as required.
- Parse source fields that encode multiple values, conditions, or list members and expand them into the required row-wise output structure.
- Produce one destination row per required emitted result, copying base fields and leaving unspecified extracted fields blank.
- Preserve source-driven ordering or destination-template ordering when possible.
- Create or locate the destination report or output block and write headers, grouped sections, or paired/list outputs only when the task calls for them.
- Use this node for summary sheets, compact report areas, and designated destination blocks for generated result lists.
- Preserve unrelated cells outside the requested output area.
- Before rewriting generated lists, reports, or expanded outputs, clear obsolete contents in the destination block that would otherwise leave stale rows, leftover formulas, or partial prior results.
- Keep headers and unrelated regions intact.
- Use the resolved output area and prior populated extent to determine what should be cleared.
- Apply only the formatting explicitly requested, such as bold text, alignment, number formats, widths, freeze panes, or copying fill/style from adjacent reference cells.
- When formatting should match a nearby column or header, clone the needed style attributes from the reference cells rather than inventing new formatting.
- Preserve existing formatting elsewhere.
- Save the modified workbook to `OUTPUT_PATH`.
- Reopen the saved workbook and inspect the explicit target cells, target range, and final fill extent directly.
- Confirm the workbook contains concrete expected-type outputs and that required cells are not left as `None` or unevaluated formulas when static results are needed.
- If verification fails, recompute from the Python logic, overwrite the target area with static values, clear stale leftover cells if necessary, resave, and recheck before finalizing.
\end{lstlisting}
\end{dialogbox2}

\end{document}

%% file: math_commands.tex
\usepackage{amsmath,amsfonts,bm}

\def\eqref#1{equation~\ref{#1}}

\def\1{\bm{1}}

\DeclareMathAlphabet{\mathsfit}{\encodingdefault}{\sfdefault}{m}{sl}
\SetMathAlphabet{\mathsfit}{bold}{\encodingdefault}{\sfdefault}{bx}{n}

